\documentclass[10pt,twocolumn,letterpaper]{article}

\usepackage[pagenumbers]{iccv} 

\usepackage{graphicx}
\usepackage{caption}
\usepackage{cuted}  

\definecolor{iccvblue}{rgb}{0.21,0.49,0.74}
\usepackage[pagebackref,breaklinks,colorlinks,allcolors=blue]{hyperref}

\def\paperID{****} 
\def\confName{ICCV}
\def\confYear{2025}

\title{\textbf{De}couple and \textbf{T}rack: Benchmarking and Improving Video Diffusion Transformers for Motion Transfer}

\author{
\centerline{
Qingyu Shi\textsuperscript{\rm 1} \
Jianzong Wu\textsuperscript{\rm 1} \
Jinbin Bai\textsuperscript{\rm 3} \
Jiangning Zhang\textsuperscript{\rm 4} \
Lu Qi\textsuperscript{\rm 5} \
Yunhai Tong\textsuperscript{\rm 1} \
Xiangtai Li\textsuperscript{\rm 2} 
}\\
\centerline{
\textsuperscript{\rm 1} PKU \quad
\textsuperscript{\rm 2} NTU \quad
\textsuperscript{\rm 3} NUS \quad
\textsuperscript{\rm 4} ZJU \quad
\textsuperscript{\rm 5} UC Merced \quad
}\\
\centerline{
Project page: \url{https://shi-qingyu.github.io/DeT.github.io}
}
}

\begin{document}
\maketitle

\captionsetup{hypcap=false}
\begin{strip}
    \vspace*{-50pt} 
    \centering
    \includegraphics[width=\linewidth]{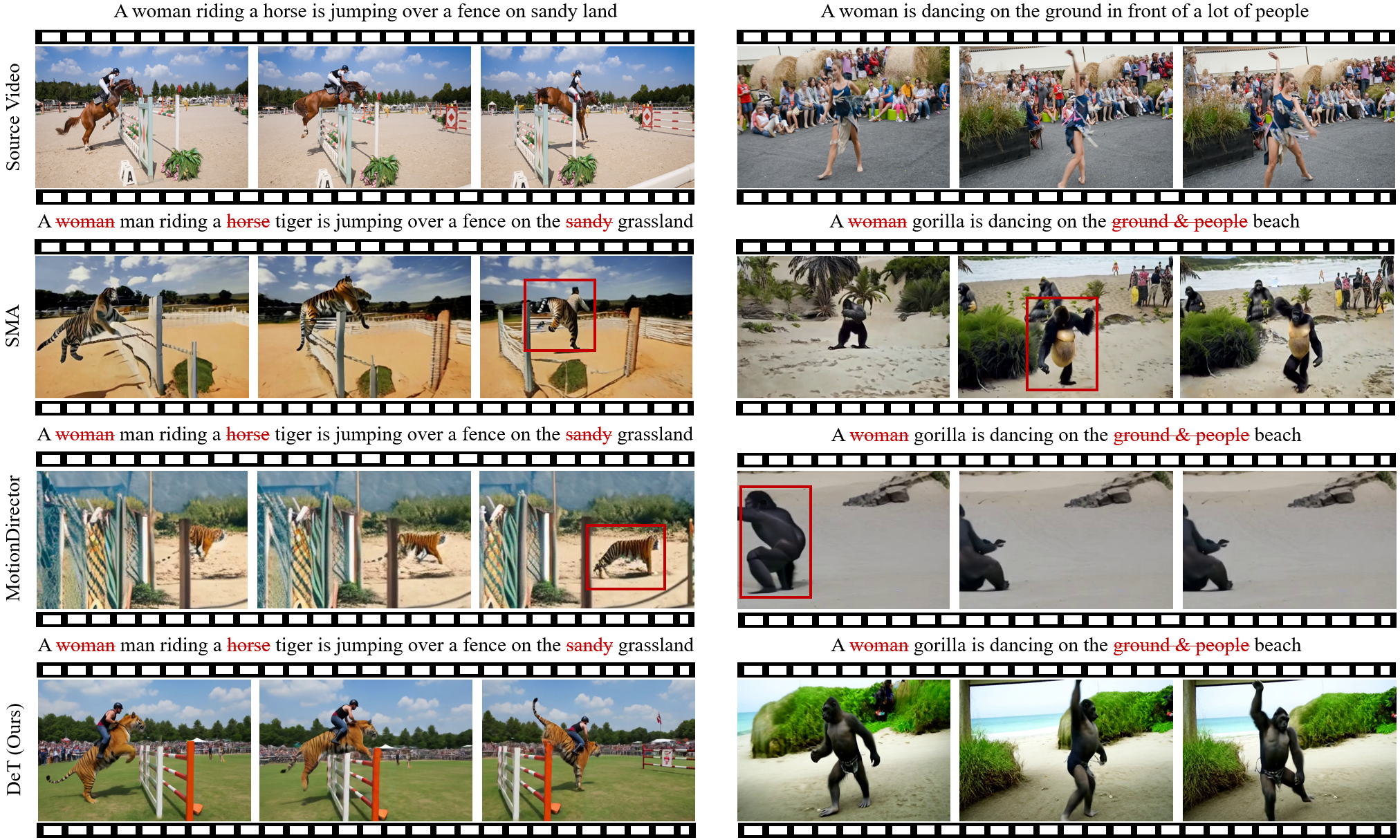}
    \captionof{figure}[teaser]{We explore the use of Diffusion Transformers (DiT) for the motion transfer task. Compared to previous works, our DeT accurately transfers motion from the source video while flexibly controlling both the foreground and background. Additionally, we highlight the worst-generated regions with red boxes.}
    \vspace{-10pt}
    \label{fig:teaser}
\end{strip}

\begin{abstract}
The motion transfer task aims to transfer motion from a source video to newly generated videos, requiring the model to decouple motion from appearance. 
Previous diffusion-based methods primarily rely on separate spatial and temporal attention mechanisms within the 3D U-Net. 
In contrast, state-of-the-art video Diffusion Transformers (DiT) models use 3D full attention, which does not explicitly separate temporal and spatial information. 
Thus, the interaction between spatial and temporal dimensions makes decoupling motion and appearance more challenging for DiT models.
In this paper, we propose DeT, a method that adapts DiT models to improve motion transfer ability. 
Our approach introduces a simple yet effective temporal kernel to smooth DiT features along the temporal dimension, facilitating the decoupling of foreground motion from background appearance. 
Meanwhile, the temporal kernel effectively captures temporal variations in DiT features, which are closely related to motion. 
Moreover, we introduce explicit supervision along dense trajectories in the latent feature space to further enhance motion consistency. 
Additionally, we present MTBench, a general and challenging benchmark for motion transfer. 
We also introduce a hybrid motion fidelity metric that considers both the global and local motion similarity. 
Therefore, our work provides a more comprehensive evaluation than previous works.
Extensive experiments on MTBench demonstrate that DeT achieves the best trade-off between motion fidelity and edit fidelity.
%
\end{abstract}    
\section{Introduction}
\label{sec:intro}
\begin{figure}[t]
  \centering
  \includegraphics[width=0.45\textwidth]{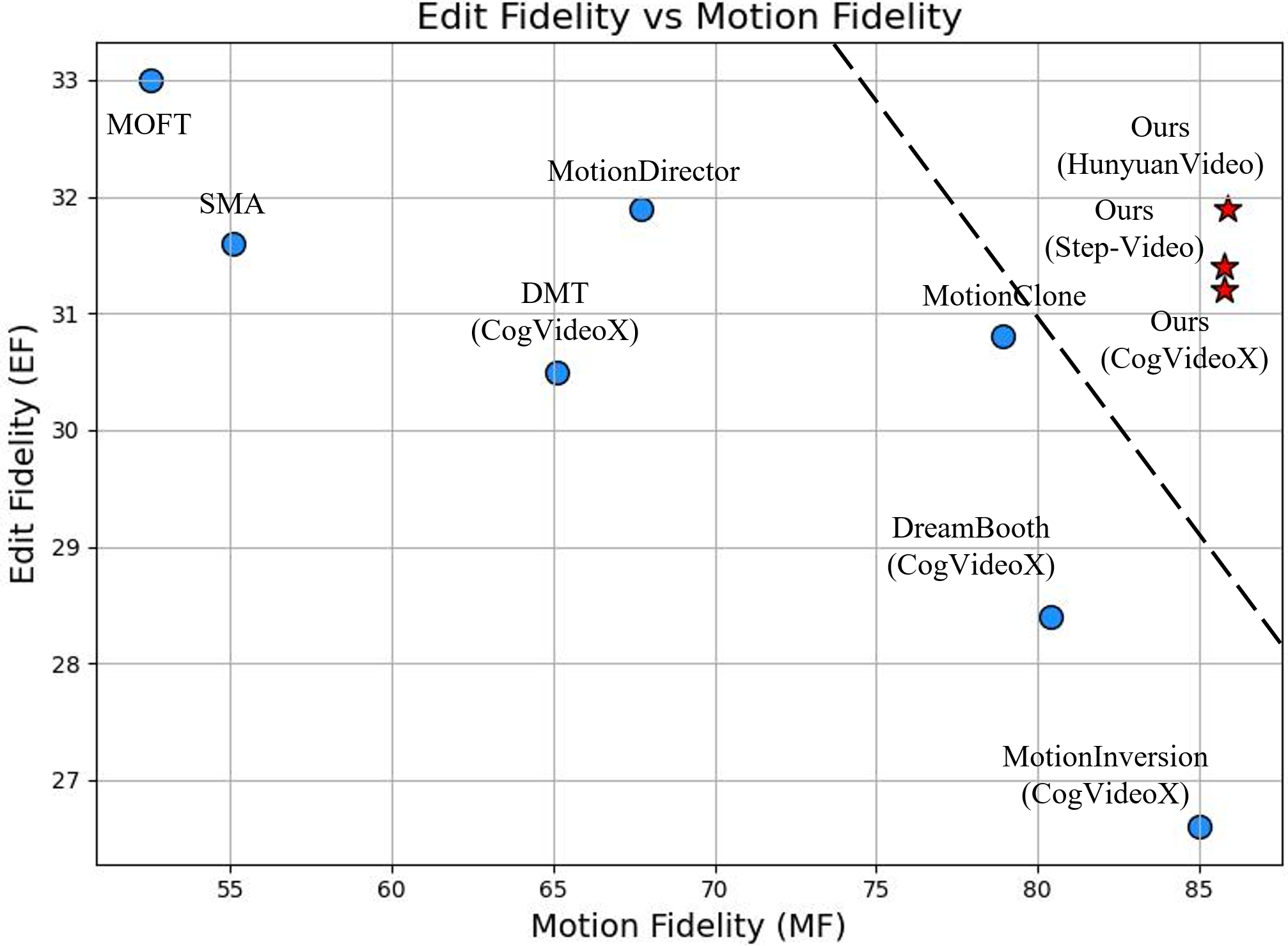}
  \vspace{-5pt}
  \caption{\textbf{Method comparision.} Our method exhibits the best balance between motion fidelity and edit fidelity.}
  \vspace{-15pt}
  \label{fig:trade_off}
\end{figure}

The advancement of foundation models for text-to-video generation has driven significant breakthroughs~\cite{animatediff, zeroscope, lavie, alignyourlatents, walt, videocrafter2}.
Notably, Diffusion Transformers (DiT) video generation models have achieved remarkable results~\cite{DiT, opensora, cogvideox, stepvideo}. 
However, generating complex motion remains a challenging, and relying solely on text prompts often fails to meet user expectations. 
To address this limitation, the motion transfer~\cite{DMT, motionclone, dreammotion, ditflow, conmo, motionprompting} seeks to generate videos that adhere to the motion in the source video under text control. 
This enables users to customize motion while retaining control over the foreground and background.

The main challenge of this task is to decouple the motion from the appearance in the source video. 
Previous methods~\cite{motiondirector, customizeavideo, vmc} mainly relied on separated temporal and spatial self-attention in 3D U-Net models~\cite{unet}, with spatial self-attention being frozen during motion learning for decoupling. 
However, these approaches are incompatible with the recent state-of-the-art DiT models~\cite{hunyuanvideo}, as their 3D full attention mechanism~\cite{cogvideox} does not explicitly separate temporal and spatial information.
The interaction of spatial and temporal dimensions in 3D attention makes the decoupling of motion and appearance more challenging.
Furthermore, in terms of the benchmark, to the best of our knowledge, previous evaluation benchmarks~\cite{DMT, motionclone} are not general enough. 
Previous evaluation benchmarks include only basic motion, such as straight-line movements~\cite{DMT}.
These benchmarks composed of simple motions fails to comprehensively evaluate the capability of motion transfer methods, limiting their practical applicability. 

In this paper, we leverage the superior performance of DiT models for the motion transfer task and introduce a general motion transfer benchmark and a hybrid metric to evaluate motion fidelity.
We first implement two existing methods, MotionInversion and DreamBooth~\cite{motioninversion, dreambooth}, which are compatible with DiT models as baselines. 
Through toy experiments shown in Fig.~\ref{fig:motion_module_comparison} (a), we observe that these baselines struggle to control the background through text. 
By visualizing the outputs of 3D full attention in Fig.~\ref{fig:motion_module_comparison}. We find that foreground and background features are difficult to distinguish, indicating that foreground motion and background appearance are not correctly decoupled. 
We argue that this is due to the temporal inconsistencies of the background feature in the denoising process, as shown in Fig.~\ref{fig:observation}.  
This makes it challenging to differentiate between foreground and background in specific frames.
To encourage the model to decouple background appearance from foreground motion, we smooth the feature in 3D full attention along the temporal dimension. 
This facilitates a more precise separation between foreground and background. 
On the other hand, to learn motion accurately, we examine the 3D attention map and find that significant attention scores appear primarily along the diagonal for adjacent frames. 
This observation motivates us to adopt a temporal 1D kernel to learn temporal information while simultaneously decoupling foreground appearance. 
The temporal kernel also effectively smooths features across time. 
By applying a trainable shared temporal kernel, we achieve motion alignment while ensuring the decoupling of both foreground and background appearance.
Considering the correspondence of optical flow in both pixel and feature spaces~\cite{raft}, we further enhance foreground motion consistency by introducing explicit supervision on latent features, named dense point tracking loss.
The dense point tracking loss aligns features along the foreground's trajectories, encouraging the model to generate consistent motion dynamics.

Furthermore, we propose a larger and more general benchmark dubbed MTBench. 
To meet the requirements for foreground-centric content and dynamic motion, MTBench is sourced from two publicly available datasets~\cite{davis, ytvos}.
We select 100 high-quality videos and leverage recent Multimodal Large Language Model~\cite{Qwen2.5-VL}, Large Language Model~\cite{qwen2.5}, and tracking model~\cite{cotracker} for video annotation. 
We generate five evaluation prompts for each source video and annotate foreground trajectories. 
The initial points for the trajectories are sampled from the mask using distance-weighted sampling. So the isolated point is more likely to be sampled. This ensures sampling within narrow but important regions such as limbs. 
An automatic clustering algorithm~\cite{silhouette} is then applied to group the trajectories, categorizing the motions into three difficulty levels based on the number of clusters. 
In addition, we also propose a hybrid motion fidelity metric.
Instead of relying solely on the similarity between local velocity~\cite{DMT} of trajectories, we introduce Fréchet distance~\cite{frechet} to measure the alignment of their global shape.
With general MTBench and the hybrid motion fidelity metric, our work offers a more comprehensive evaluation of motion transfer methods.

Building on MTBench, both qualitative and quantitative results highlight the superior performance of our method in achieving an optimal balance between motion and edit fidelity. Further ablation studies confirm the effectiveness of our proposed shared temporal kernel and dense point tracking loss.
Notably, as shown in Fig.~\ref{fig:teaser}, our approach is good at transferring the motion between cross-category foregrounds, resulting in a more generalized motion transfer method.
To summarize, our key contributions include:
\begin{itemize}
    \item An effective method that decouples and learns motion patterns simultaneously.
    \item We introduce a dense point tracking loss in the latent space to enhance foreground motion consistency. 
    \item A challenging and comprehensive benchmark for the motion transfer task, dubbed MTBench.
    \item A hybrid metric for evaluating motion fidelity.
    \item We achieve the best trade-off between motion and edit fidelity on MTBench as shown in Fig.~\ref{fig:trade_off}.
\end{itemize}

\section{Related Work}
\label{sec:related_work}

\noindent
\textbf{Text-to-video diffusion models.}
Building on recent text-to-image (T2I) diffusion models~\cite{ldm, sdxl, pixart, bai2024meissonic, sd3}, early approaches~\cite{animatediff, videocrafter2} extended pre-trained T2I models to generate dynamic videos. 
These methods typically incorporate temporal self-attention within the U-Net~\cite{unet} architecture to ensure motion dynamics. 
However, due to the per-patch computation, the generated videos often exhibit temporal inconsistencies in motion. 
Recently, the DiT models~\cite{cogvideox, ltxvideo, hunyuanvideo, stepvideo} introduced 3D Full Attention, and through end-to-end training on large-scale video datasets~\cite{webvid10M, panda70M}, it is capable of generating temporally consistent and high-quality videos.
However, generating motion solely from text prompts remains challenging, particularly when it comes to meeting user demands for complex motions.
In this paper, we propose a novel method that enables users to generate specific motions within a source video under text control.


\noindent
\textbf{Motion transfer.}
This task aims to transfer the motion from a source video to a generated video. 
Unlike video editing tasks~\cite{tokenflow, videop2p, sdedit, tuneavideo, deltadit, bai2023integrating, feng2024item, bai2024humanedit}, motion transfer focuses solely on the motion pattern in the source video, allowing for more flexible control of the foreground and background in the generated video through text. 
The main challenge in motion transfer is decoupling motion from appearance in the source video, enabling effective learning of the motion while avoiding memorizing appearance. 
Previous methods~\cite{motiondirector, vmc, motioninversion, customizeavideo, sma} relied on a U-Net architecture with a decoupled spatial and temporal self-attention mechanism. They fine-tune the parameters in the temporal self-attention only to ensure the motion is learned without overfitting. 
However, the state-of-the-art DiT models~\cite{hunyuanvideo} use 3D full attention without explicit spatial temporal decoupling, making previous approaches incompatible with DiT models. 
We propose the shared temporal kernel and dense point tracking loss to decouple and track motion in the source video, enabling DiT models to effectively transfer it to newly generated video. 
\section{Method}
\label{sec:method}

\begin{figure}[t]
  \centering
  \includegraphics[width=0.47\textwidth]{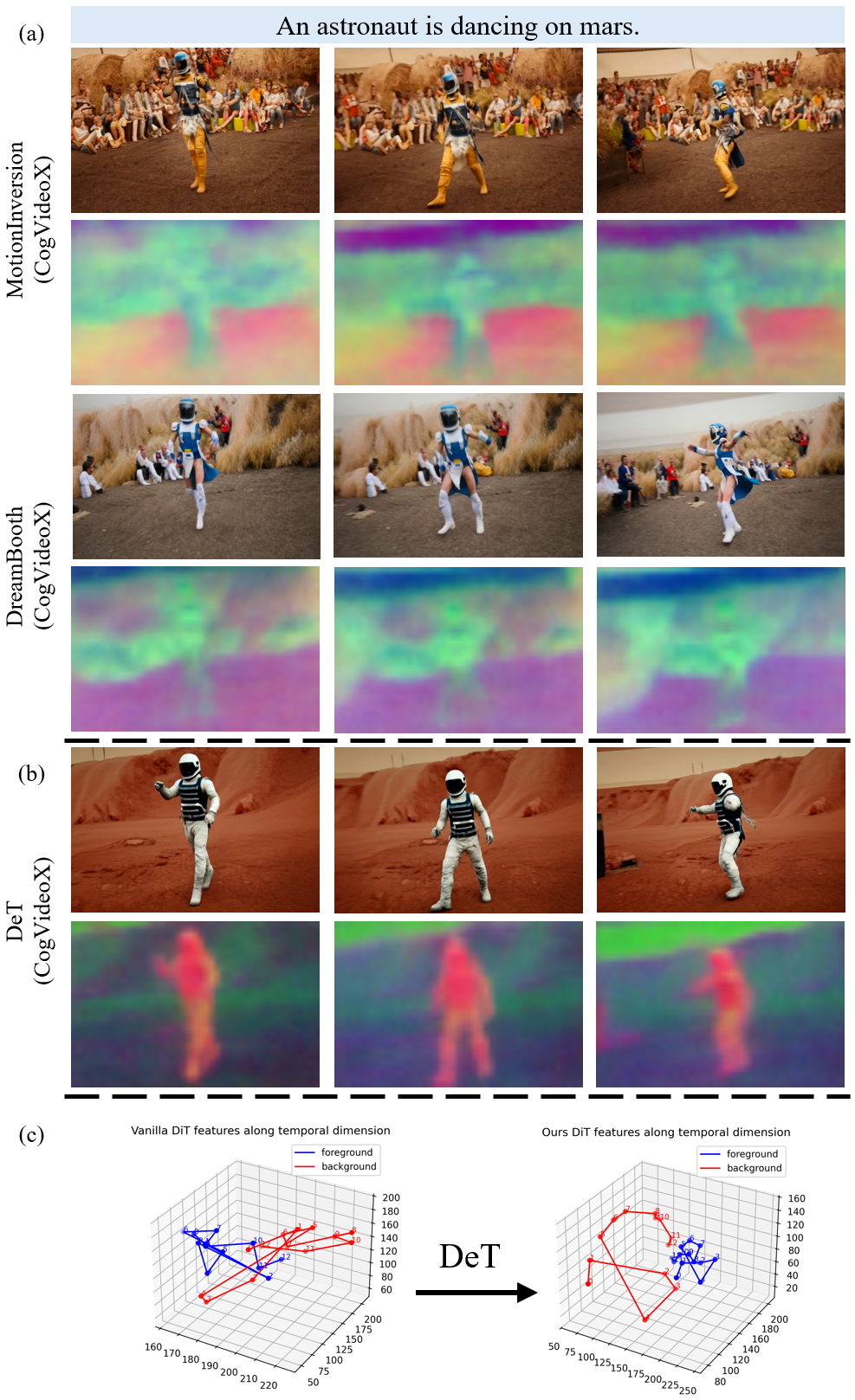}
  \vspace{-5pt}
  \caption{\textbf{Comparison of DiT features.} The source video is shown in Fig.~\ref{fig:observation}. We use CogVideoX-5B as the default model, extracting and averaging DiT features from the 4th block across all denoising steps. (a) and (b) is the baseline and our method's generated frames and DiT features, respectively. (c) Illustrate the DiT feature along the temporal dimension.}
  \vspace{-12pt}
  \label{fig:motion_module_comparison}
\end{figure}

\noindent
\textbf{Notation.} We follow DDPM~\cite{DDPM}: $t$ is the diffusion step, $\epsilon$ Gaussian noise, $\bar{\alpha}_t$ the cumulative noise schedule, and $\epsilon_\theta(\cdot)$ the noise predictor.

Given a source video $\mathcal{S}$ and a text prompt $P$, our goal is to generate a new video $\mathcal{J}$ that retains the overall motion of $\mathcal{S}$ while aligning with $P$. 
In the following section, we first introduce two baseline approaches and examine their weaknesses, and then we present our method to decouple motion from appearance in $\mathcal{S}$.

\subsection{Motivations}

\begin{figure}[t]
  \centering
  \includegraphics[width=0.47\textwidth]{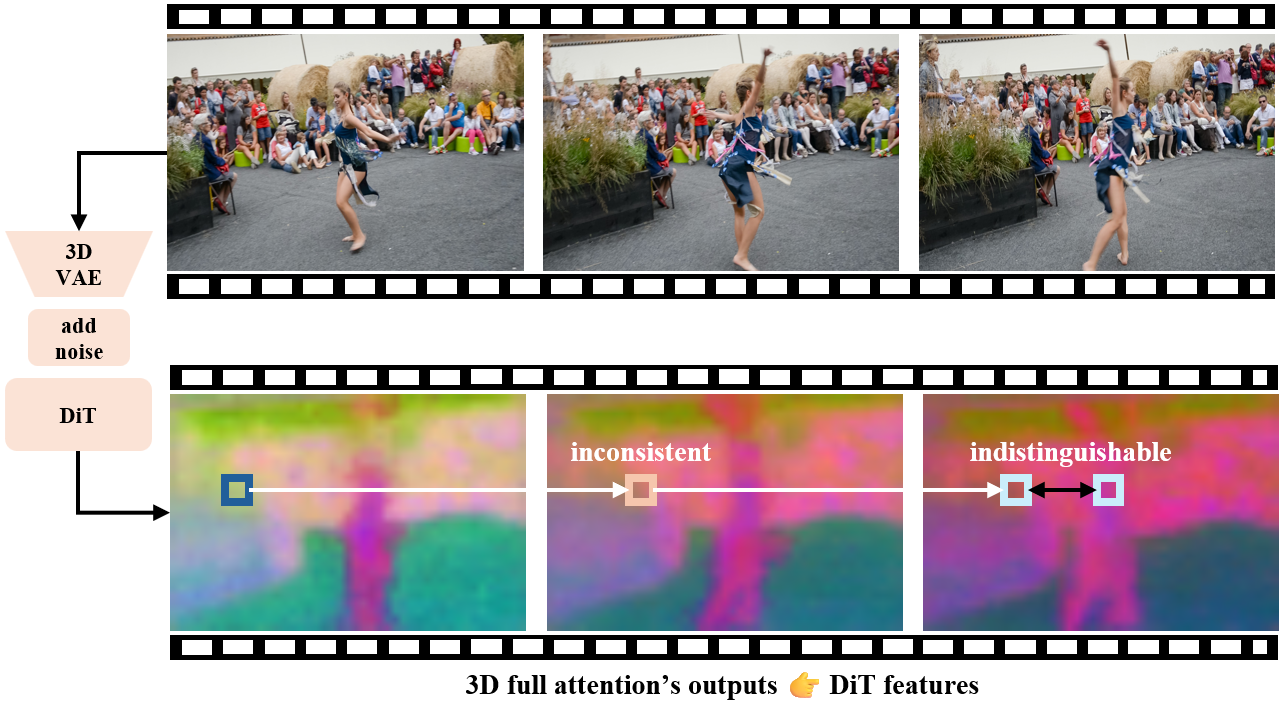}
  \vspace{-5pt}
  \caption{\textbf{The visualization of DiT features.} We attribute the failure to decouple background appearance to the difficulty of distinguishing foreground and background features in certain frames. As a result, background appearance becomes entangled with foreground motion during training.}
  \vspace{-12pt}
  \label{fig:observation}
\end{figure}

\begin{figure}[t]
  \centering
  \includegraphics[width=0.47\textwidth]{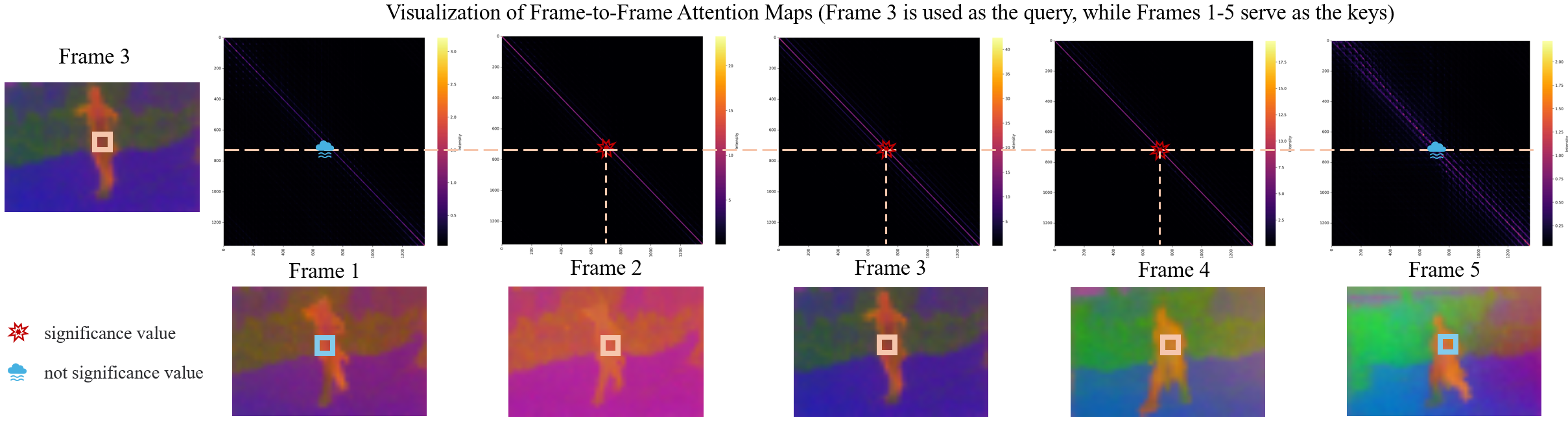}
  \vspace{-5pt}
  \caption{\textbf{The visualization of frame-to-frame attention map.} We mark significant and non-significant values in the figure. Significant values appear only along the diagonal of adjacent frames, suggesting that capturing DiT feature changes over time requires only a local temporal receptive field, not a spatial one.}
  \vspace{-15pt}
  \label{fig:attention_map}
\end{figure}

Since previous works only fine-tune temporal attention in 3D U-Net~\cite{vmc, sma, customizeavideo, dreamvideo, dreammotion}, they are not compatible with DiT models. 
As baselines, we implement the embedding-based method MotionInversion~\cite{motioninversion} and the fully fine-tuned method DreamBooth~\cite{dreambooth} on DiT models. 
As shown in Fig.~\ref{fig:motion_module_comparison} (a), neither method effectively controls the appearance of the foreground and background using text. Moreover, the entangled motion and appearance information further complicates motion learning.

To investigate this issue and dig out reasons, we extract the output of 3D full attention, referred to as the DiT feature $\mathcal{I} \in \mathbb{R}^{thw\times c}$, where $t, h, w$ are compressed frames, height, and width respectively. 
We visualize $\mathcal{I}$ using Principle Component Analysis (PCA) in Fig.~\ref{fig:motion_module_comparison} (a). The results reveal that foreground and background features are difficult to distinguish, suggesting that background appearance and foreground motion are not properly decoupled. 
Consequently, when generating foreground motion, the background appearance becomes entangled with it.

To address this issue, we analyze the DiT feature $\mathcal{I}$ during the denoising process. 
As shown in Fig.~\ref{fig:observation}, we observe that temporal inconsistencies in \textit{background features} make distinguishing between foreground and background challenging. 
To separate foreground and background, an ideal approach is to smooth foreground and background features along the temporal dimension.
Additionally, in order to accurately learn the motion pattern, we analyze the 3D attention map.
The Fig.~\ref{fig:attention_map} shows that strong diagonal attention scores appear only in adjacent frames. This suggests that modeling temporal changes in DiT features does not depend on the spatial receptive field and should instead focus on \textit{local temporal} information. 
The reduction of spatial dimension also encourages the model to decouple the foreground appearance.
Based on the above analysis, leveraging smoothness and temporal learning enables the decoupling of both foreground and background appearance, leading to successful motion transfer.

\subsection{DeT}

\begin{figure}[t]
  \centering
  \includegraphics[width=0.45\textwidth]{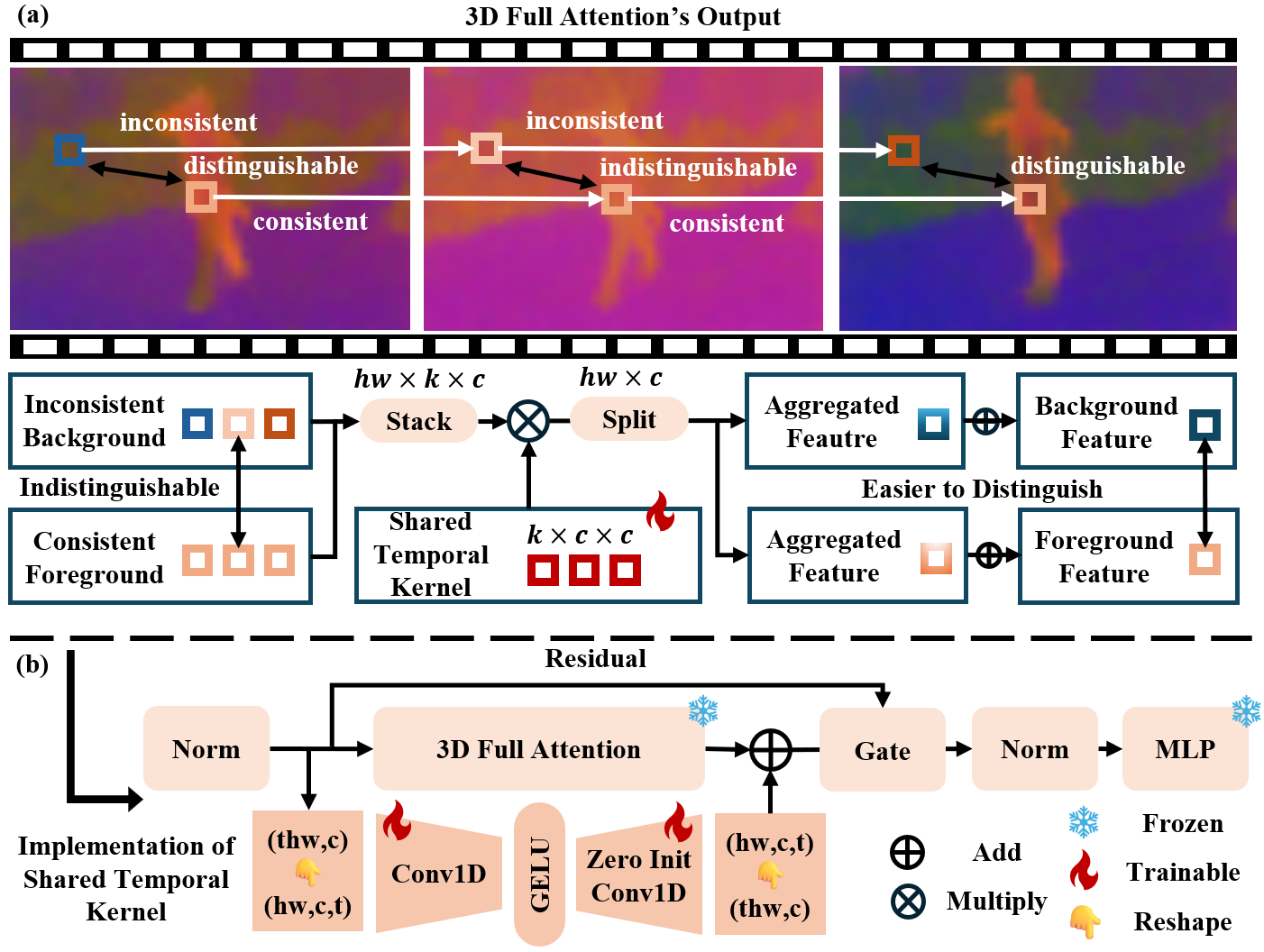}
  \vspace{-5pt}
  \caption{\textbf{Overview of the shared temporal kernel.} Figure (a) illustrates the smoothing mechanism of the shared temporal kernel, which aggregates DiT features along the temporal dimension. Figure (b) presents its implementation using temporal 1D convolution with a GELU activation function.}
  \vspace{-12pt}
  \label{fig:motion_module}
\end{figure}

\begin{table*}[t]
\centering
\scriptsize
\setlength{\tabcolsep}{6pt}
\scalebox{0.90}{
\begin{tabular}{lccccc}
\toprule
                 & DMT~\cite{DMT} & MotionDirector~\cite{motiondirector} & MotionClone~\cite{motionclone} & SMA~\cite{sma} & MTBench \\
\hline
Num Videos   & 21 & 76  & 40  & 30  & \textbf{100} \\
Num Prompts  & 56 & 228 & Unknown & Unknown & \textbf{500} \\
Benchmark Source & DAVIS~\cite{davis} \& Web Source & LOVEU-TGVE~\cite{loveu} & DAVIS~\cite{davis} \& Web Source & DAVIS~\cite{davis} \& WebVid10M~\cite{webvid10M} & DAVIS~\cite{davis} \& YouTube-VOS~\cite{ytvos} \\
Domain           & Animal \& Vehicle   & Human \& Animal \& Vehicle & Human \& Animal \& Vehicle & Human \& Animal \& Vehicle & Human \& Animal \& Vehicle \\
Difficulty & Easy & Easy \& Medium & Easy \& Medium & Easy \& Medium & Easy \& Medium \& Hard \\
Num Clusters      & 2.57  & 4.72  & 4.13  & 3.91  & 6.31 \\
\bottomrule
\end{tabular}
}
\vspace{-5pt}
\caption{\textbf{Benchmark Comparison.} We present a table summarizing the scale and difficulty of previous benchmarks. The bottom row is the average number of clusters.}
\vspace{-15pt}
\label{tab:benchmark_comparison}
\end{table*}

\begin{figure}[t]
  \centering
  \includegraphics[width=0.44\textwidth]{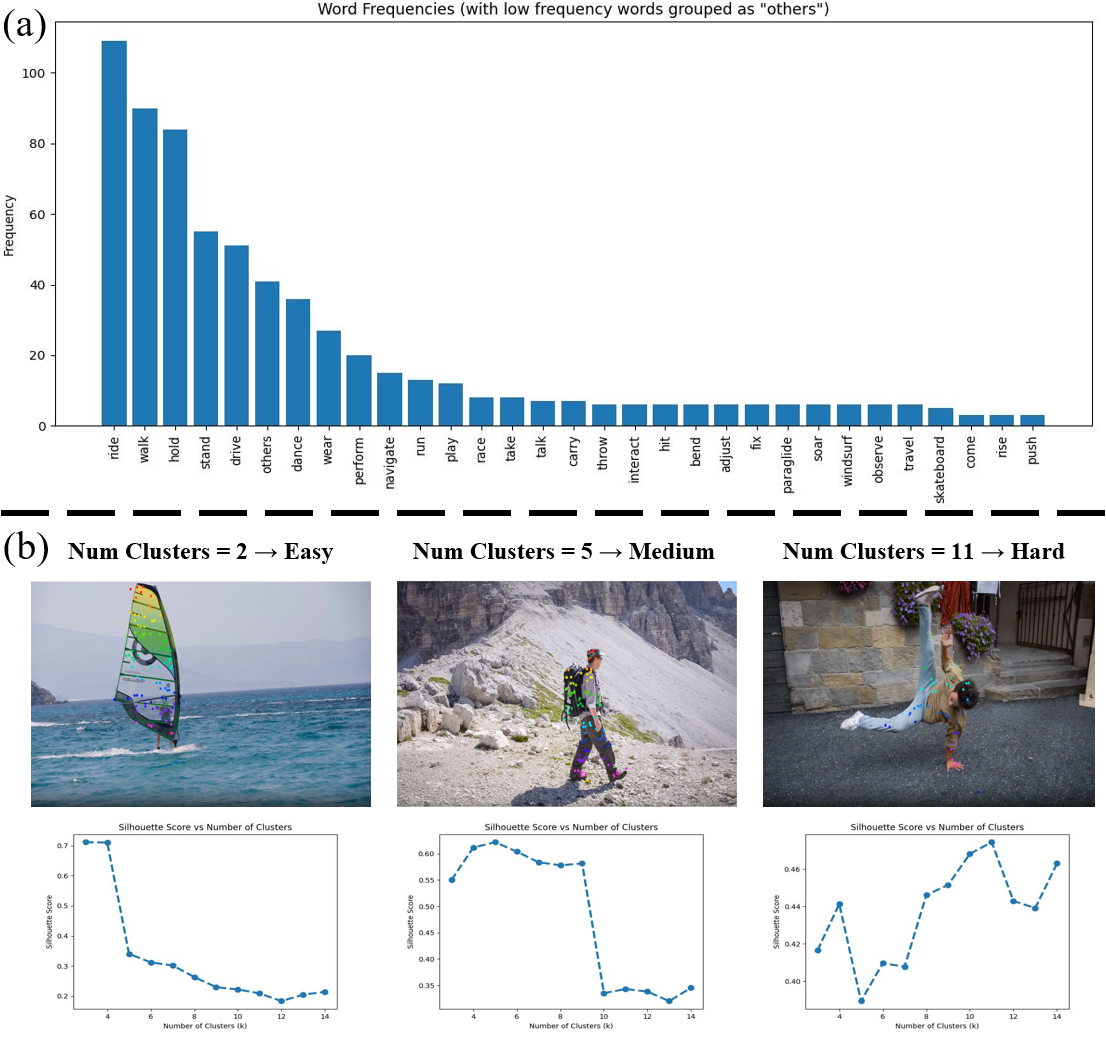}
  \vspace{-5pt}
  \caption{\textbf{Overview of MTBench.} (a) illustrates the motion frequency distribution in evaluation prompts. (b) illustrates difficulty levels based on the number of trajectory clusters: 1-3 as easy, 4-6 as medium, and 7+ as hard. The average number of clusters for the easy, medium, and hard levels are 2.7, 5.8, and 10.4, respectively.}
  \vspace{-15pt}
  \label{fig:word_cloud}
\end{figure}

\noindent
\textbf{Shared temporal kernel.}
Building on the motivation, we first apply temporal smoothness to the reshaped DiT features \( \mathcal{I}' \in \mathbb{R}^{hw \times t \times c} \). Specifically, we use a temporal kernel \( \mathcal{K}^t \in \mathbb{R}^{k \times c \times c} \) to perform convolution along the temporal dimension. From a manifold learning perspective, the temporal kernel functions as an equivalent Laplacian smoothing operator along the temporal axis:
\begin{equation}
    \hat{\mathcal{I}}_{xy,\,i} = \mathcal{I}'_{xy,\,i} + \sum_{j=-\frac{k-1}{2}}^{\frac{k-1}{2}} \mathcal{I}'_{xy,\,i+j} \times \mathcal{K}^{\text{t}}_{j}
\end{equation}
, where $xy$ and $i$ represent the spatial and temporal coordinates, respectively. 
After applying smoothness, the model can utilize the refined temporal trajectories to achieve more reliable differentiation.

Furthermore, we find that the temporal kernel also serves as an ideal model for motion learning. As discussed above, motion learning primarily depends on local temporal information rather than a spatial receptive field. The shared temporal kernel along the spatial dimension effectively meets both requirements. Thus, we incorporate it into all DiT blocks. Specifically, as shown in Fig.~\ref{fig:motion_module}, to enhance efficiency and reduce memory consumption, we adopt a down-and-up architecture $\mathcal{K}^{\text{t}}_{down} \in \mathbb{R}^{k\times c\times m},\, \mathcal{K}^{\text{t}}_{up} \in \mathbb{R}^{k\times m \times c}$ with a GELU activation $\sigma$ function in the middle:
\begin{equation}
    \tilde{\mathcal{I}} = \mathcal{K}^{\text{t}}_{up} * \sigma\,(\mathcal{K}^{\text{t}}_{down} * \mathcal{I}) + \text{Attention}(\mathcal{I}, \mathcal{E}_{text})
\end{equation}
where $\mathcal{E}_{text}$ represents the embedding of the text prompt and $*$ is the convolution operation. As a toy experiment on DiT, Fig.~\ref{fig:motion_module_comparison} (b) shows that the generated video follows the motion pattern of the source video while maintaining strong text controllability. Furthermore, Fig.~\ref{fig:motion_module_comparison} (c) demonstrates that our shared temporal kernel effectively enhances the model's ability to distinguish between foreground and background DiT features.

\noindent
\textbf{Dense point tracking loss.}
Inspired by our finding that foreground DiT features remain consistent over time, we introduce more explicit supervision on foreground DiT features along the temporal dimension to enhance the foreground motion consistency. 

Specifically, we utilize CoTracker~\cite{cotracker} to track the foreground in the source video, generating a set of trajectories $\mathcal{T}\in \mathbb{R}^{N\times T\times 2}$, where $N$ denotes the number of trajectories. 
Additionally, CoTracker provides the visibility of each point in the form of $\mathcal{V}\in \{0, 1\}^{N\times T\times 1}$, where 1 indicates visibility, and 0 denotes invisibility. 
During the training process on source video $\mathcal{S}$, we calculate dense point tracking loss on the predicted latent feature $\hat{\mathcal{E}}(\mathcal{S}) \in \mathbb{R}^{t\times h\times w\times c}$:
\begin{align}
  \hat{\mathcal{E}}(\mathcal{S}) &= \sqrt{\bar{\alpha}_t}\,\Bigl(\sqrt{\bar{\alpha}_t}\, \mathcal{E}(\mathcal{S}) + \sqrt{1-\bar{\alpha}_t}\,\epsilon\Bigr) - \sqrt{1-\bar{\alpha}_t}\,v
\end{align}
where $v = \epsilon_{\theta}\Bigl(\sqrt{\bar{\alpha}_t}\, \mathcal{E}(\mathcal{S}) + \sqrt{1-\bar{\alpha}_t}\,\epsilon \Bigr)$. We align the features along the trajectories $\mathcal{T}$ using the L2 distance. Additionally, to account for occlusions where parts of the trajectory may not be visible, we compute the dense point tracking loss only on visible points by applying $\mathcal{V}$ as a mask:
\begin{align}
  \mathcal{L}_{TL} &= \left\| \min\Bigl( \mathcal{V}(t+1),\, \mathcal{V}(t) \Bigr) \times \mathcal{L} \right\|_2^2
\end{align}
where $  \mathcal{L} = \hat{\mathcal{E}}(\mathcal{S})\bigl[\mathcal{T}(t+1)\bigr] - \hat{\mathcal{E}}(\mathcal{S})\bigl[\mathcal{T}(t)\bigr]$. The final loss is a combination of denoising loss (DL) and dense point tracking loss (TL), weighted by $\lambda_{DL}$ and $\lambda_{TL}$:
\begin{equation}
    \mathcal{L} = \lambda_{DL}\, \mathcal{L}_{DL} + \lambda_{TL}\,\mathcal{L}_{TL}
\end{equation}

\begin{figure*}[t]
  \centering
  \includegraphics[width=0.97\textwidth]{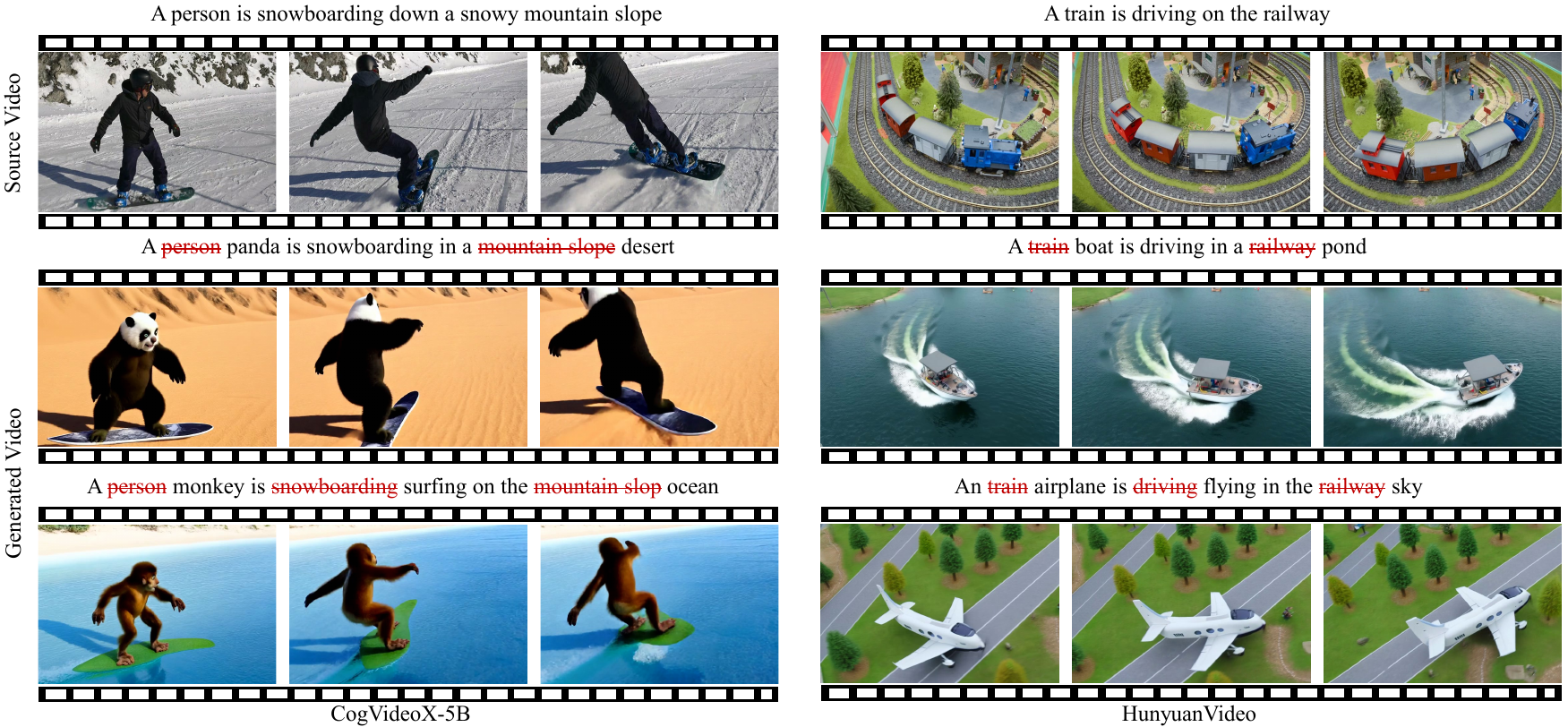}
  \vspace{-5pt}
  \caption{\textbf{Qualitative results of DeT.} The left column of videos is generated by CogVideoX-5B, while the right column is generated by HunyuanVideo. DeT can transfer motion from easy to hard difficulty levels. The generated video adheres to the motion pattern of the source video while allowing flexible foreground and background control through text.}
  \vspace{-10pt}
  \label{fig:qualitative_results}
\end{figure*}

\vspace{-5pt}
\begin{figure*}[t]
  \centering
  \includegraphics[width=0.97\textwidth]{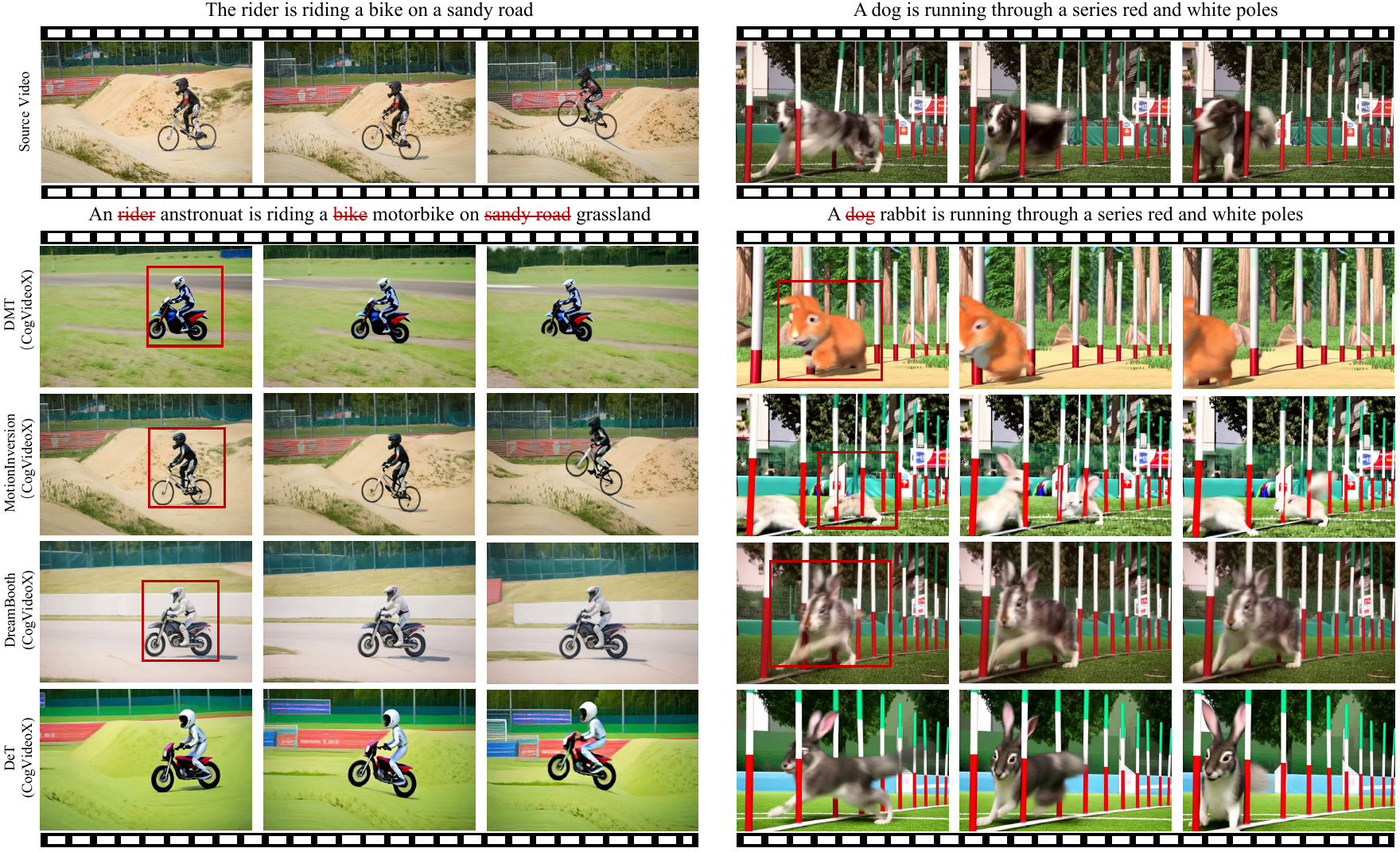}
  \vspace{-5pt}
  \caption{\textbf{Qualitative comparison.} We compare DeT with other state-of-the-art motion transfer methods. We specify the base models used in parentheses. DeT demonstrates a clear advantage in both motion and edit fidelity.}
  \vspace{-10pt}
  \label{fig:qualitative_comparison}
\end{figure*}

\subsection{MTBench}
\subsubsection{Data Collection and Annotation}
As shown in Tab.~\ref{tab:benchmark_comparison}, existing benchmarks are limited in either scale (number of videos and prompts) or difficulty. To enable a more comprehensive evaluation of motion transfer,
we construct a new benchmark based on DAVIS~\cite{davis} and YouTube-VOS~\cite{ytvos}, consisting of 100 high-quality videos and 500 evaluation prompts.
Captions are first generated using Qwen2.5-VL-7B~\cite{Qwen2.5-VL}, and five evaluation prompts per video are derived with Qwen2.5-14B~\cite{qwen2.5} by swapping foreground and background while preserving the verb. Subject masks and trajectories are annotated with SAM~\cite{sam} and CoTracker~\cite{cotracker}. We then perform K-means clustering of trajectories based on the silhouette coefficient~\cite{silhouette} to categorize motion into easy, medium, and hard levels by the number of clusters.
Figure~\ref{fig:word_cloud} (a) and (b) illustrate the motion distribution and difficulty division, while Tab.~\ref{tab:benchmark_comparison} highlights the broader motion diversity of our benchmark compared to prior works.

\subsubsection{Metric}
\noindent
\textbf{Hybrid motion fidelity (motion fidelity, for short).}
We introduce Fréchet distance $d_F$~\cite{frechet, FMD} to compare the shape differences between trajectories $\mathcal{T}_i,\,\mathcal{T}_j\in \mathbb{R}^{N\times T\times 2}$. It considers both the spatial arrangement and the order of points. This ensures that the transferred motion globally adheres to the source motion. We also use cosine similarity~\cite{DMT} to compare the local velocity direction $\Delta \mathcal{T}(t) = \mathcal{T}(t+1) - \mathcal{T}(t)$. Specifically, we compute motion fidelity score $\mathcal{M}$ through the formulation:
\[
\mathcal{M}(\mathcal{T}_i,\mathcal{T}_j)=\frac{1}{N}\sum_{n=1}^{N}\Bigl[\alpha\,e^{-d_F(\mathcal{T}_i^n,\mathcal{T}_j^n)}+(1-\alpha)\,\bar{c}_n\Bigr],
\]
where $\bar{c}_n=\frac{1}{T-1}\sum_{t=1}^{T-1}\cos\Bigl(\Delta\mathcal{T}_i^n(t),\,\Delta\mathcal{T}_j^n(t)\Bigr).$ and $\alpha \in [0, 1]$ is a weighting parameter that balances the shape and velocity similarity. 
We set $\alpha=0.5$ in our experiments.

\noindent
\textbf{Edit fidelity.}
Following previous works~\cite{clipscore, dreambooth, vbench}. We use CLIP~\cite{clip} to measure the similarity between each frame and the target text prompt, reporting the average score.

\noindent
\textbf{Temporal consistency.}
The motion learning process may cause temporal flickering. We compute the cosine similarity between the DINO~\cite{dinov2} features of consecutive frames to measure the temporal consistency of the generated videos.
\section{Experiment}
\label{sec:exp}

\begin{table*}[t]
\centering
\scalebox{0.95}{
\begin{tabular}{lccc|ccc|ccc|ccc}
\toprule
 & \multicolumn{3}{c}{All} & \multicolumn{3}{c}{Hard} & \multicolumn{3}{c}{Medium} & \multicolumn{3}{c}{Easy} \\
\cmidrule(lr){2-4} \cmidrule(lr){5-7} \cmidrule(lr){8-10} \cmidrule(lr){11-13}
Method & EF & TC & MF & EF & TC & MF & EF & TC & MF & EF & TC & MF \\
\midrule
MotionDirector                & \underline{31.9} & \textbf{91.7} & 67.7 & 32.3 & \underline{88.3} & 67.9 & \underline{32.5} & \textbf{93.4} & 66.8 & \underline{30.8} & \textbf{93.4} & 68.3 \\
SMA                           & 31.6 & 82.9 & 55.1 & \underline{32.9} & 78.1 & 44.1 & 32.1 & 84.7 & 52.1 & 29.9 & 85.8 & 69.2 \\
\midrule
MotionClone                   & 30.8 & 80.9 & 78.9 & 30.5 & 74.4 & 72.5 & 31.9 & 83.4 & 79.2 & 30.0 & 84.8 & \textbf{85.0} \\
MOFT                          & \textbf{33.0} & \underline{91.1} & 52.5 & \textbf{33.8} & \textbf{89.7} & 48.0 & \textbf{33.2} & \underline{92.1} & 53.2 & \textbf{31.9} & 91.4 & 56.3 \\
\midrule
DMT (CogVideoX)               & 30.5 & 87.6 & 65.1 & 31.6 & 84.1 & 64.1 & 31.8 & 89.0 & 64.6 & 28.2 & 89.6 & 66.7 \\
DreamBooth (CogVideoX)        & 28.4 & 85.6 & 80.4 & 27.5 & 83.2 & 76.9 & 27.9 & 84.7 & 80.1 & 29.9 & 88.9 & 84.2 \\
MotionInversion (CogVideoX)   & 26.6 & 85.4 & \underline{85.0} & 27.2 & 80.8 & \textbf{80.7} & 27.2 & 84.9 & \underline{90.8} & 25.5 & 89.4 & 84.6 \\
Ours (CogVideoX)              & 31.2 & 89.6 & \textbf{85.8} & 31.8 & 84.7 & \underline{80.3} & 31.1 & 91.5 & \textbf{92.2} & 30.7 & \underline{92.7} & \underline{84.8} \\
\midrule
Ours (HunyuanVideo)           & 31.9 & 91.9 & 85.9 & 31.1 & 89.7 & 80.1 & 32.3 & 93.1 & 92.2 & 31.1 & 92.9 & 85.4 \\
Ours (Step-Video-T2V)         & 31.4 & 91.6 & 85.8 & 30.9 & 87.6 & 80.5 & 31.7 & 93.5 & 91.9 & 31.6 & 92.7 & 85.0 \\
\bottomrule
\end{tabular}}
\vspace{-5pt}
\caption{\textbf{Quantitative results on MTBench.}  EF: Edit Fidelity, TC: Temporal Consistency, MF: Motion Fidelity.  For each column, the best score is shown in bold and the second‐best is underlined; the last two rows are not included in the ranking for a fair comparison.}
\vspace{-10pt}
\label{tab:quantitative_results}
\end{table*}

\begin{figure}[t]
  \centering
  \includegraphics[width=0.47\textwidth]{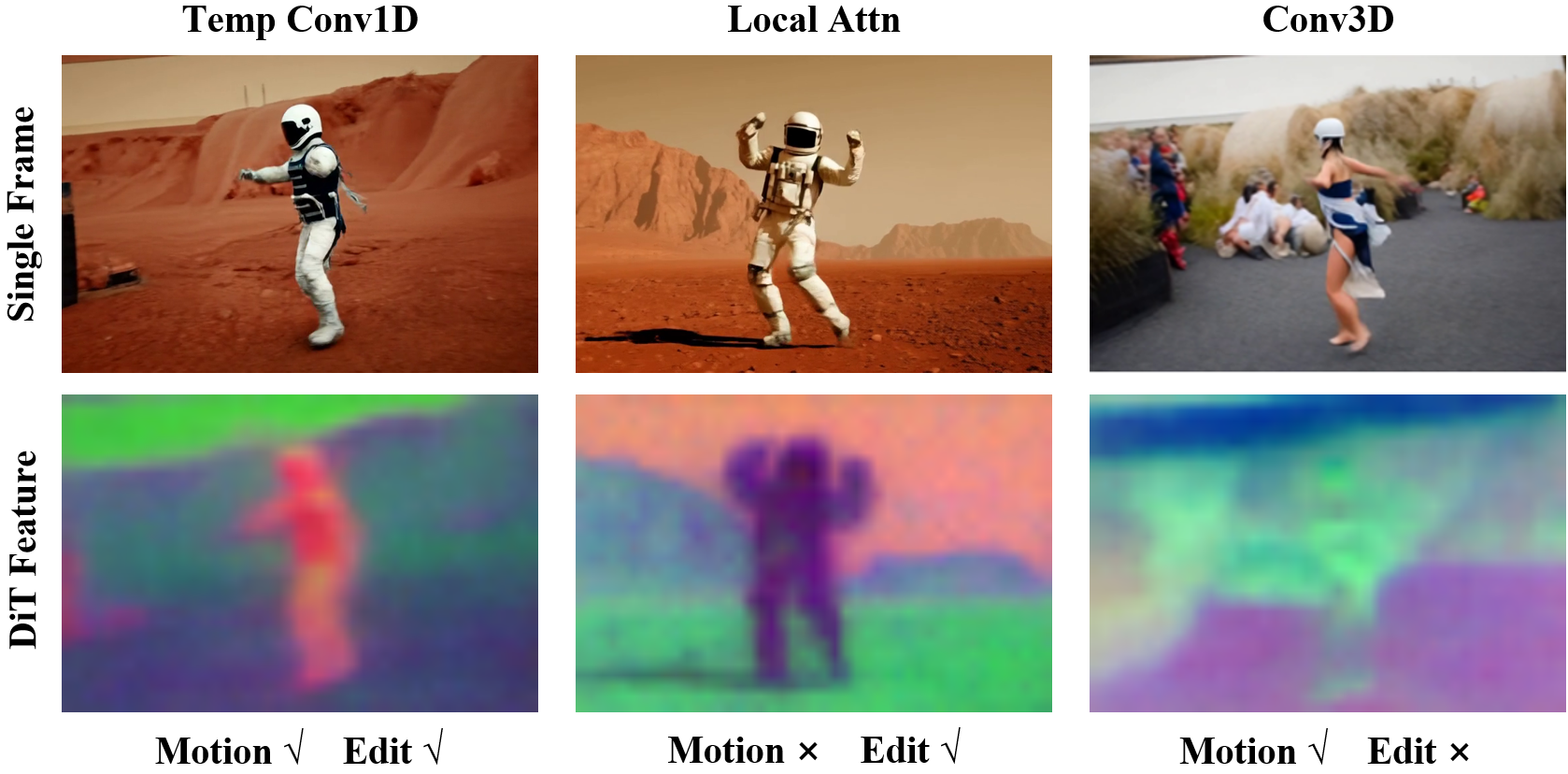}
  \vspace{-10pt}
  \caption{\textbf{Ablation study on shared temporal kernel.} Temporal kernel achieve both motion fidelity and edit fidelity.}
  \vspace{-10pt}
  \label{fig:temp_conv1d_ablation}
\end{figure}

\begin{figure}[t]
  \centering
  \includegraphics[width=0.47\textwidth]{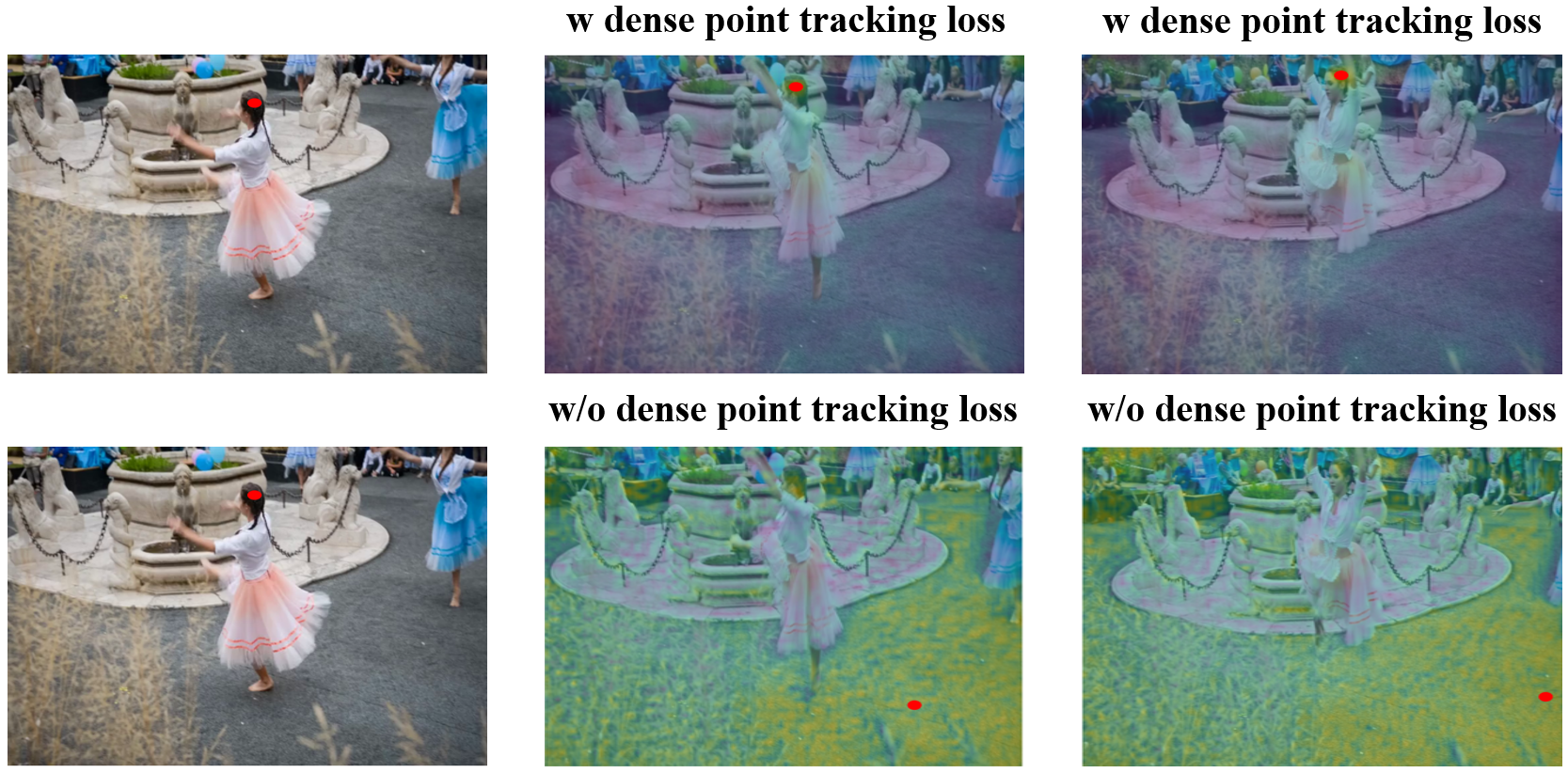}
  \vspace{-10pt}
  \caption{\textbf{Ablation study on dense point tracking loss.} We visualize the cosine similarity across video frames.}
  \vspace{-10pt}
  \label{fig:tracking_loss_ablation_2}
\end{figure}

\begin{figure}[t]
  \centering
  \includegraphics[width=0.47\textwidth]{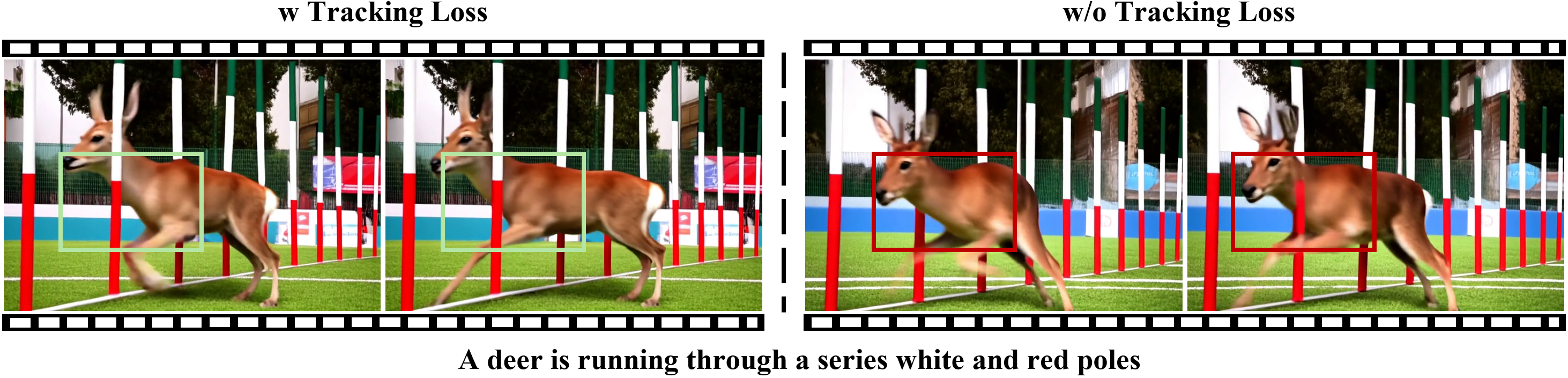}
  \vspace{-10pt}
  \caption{\textbf{Ablation study on dense point tracking loss.} Improvements are highlighted using green and red boxes.}
  \vspace{-10pt}
  \label{fig:tracking_loss_ablation}
\end{figure}

\subsection{Experimental Setup}
\noindent
\textbf{Implementation details.}
We apply DeT on three pre-trained DiT models—CogVideoX-5B~\cite{cogvideox}, HunyuanVideo~\cite{hunyuanvideo}, and StepVideo-T2V~\cite{stepvideo} to demonstrate its generalization capabilities. We just provide implementation details for CogVideoX as the base model in the main paper. For HunyuanVideo and Step-Video-T2V, please refer to the Supplementary Material (SM). We train DeT for 500 steps on a single source video using the AdamW optimizer~\cite{adam} with a learning rate of 1e-5 and a weight decay of 1e-2. The loss weight parameters $\lambda_{DL}$ and $\lambda_{TL}$ is set to 1.0 and 0.1. The mid-dimension is set to 128 for the shared temporal kernel, and the kernel size is configured to 3. During training, we integrate the shared temporal kernel into all DiT blocks. During inference, we remove the shared temporal kernel from the last 27 blocks (65\%). This operation improves edit fidelity, as explained in the SM. The injected shared temporal kernel efficiently transfers motion from the source video to the generated video under text control. We perform 50-step denoising using the DDIM scheduler and set the classifier-free guidance scale to 6.0. The generated videos are $49\times480\times720$. The training process finishes in around 1 hour in a single NVIDIA A100 GPU. 

\noindent
\textbf{Baselines.}
We compare our method against state-of-the-art motion transfer approaches, including MotionDirector~\cite{motiondirector}, SMA~\cite{sma}, MOFT~\cite{MOFT}, and MotionClone~\cite{motionclone}. Since prior methods rely on 3D U-Nets and cannot be directly applied to DiT, we additionally implement DMT~\cite{DMT}, MotionInversion~\cite{motioninversion}, and DreamBooth LoRA~\cite{dreambooth} on DiT for a fair comparison.

\subsection{Qualitative Results}
As illustrated in Fig.~\ref{fig:qualitative_results}, our method faithfully transfers motion from the source video without overfitting to its appearance, enabling flexible text control over both foreground and background. It further supports cross‑category motion transfer, e.g., from a human to a panda or a train to a boat.
Fig.~\ref{fig:teaser} and~\ref{fig:qualitative_comparison} show that U‑Net‑based methods~\cite{motiondirector, sma} struggle with motion fidelity and exhibit weak text‑editing fidelity. When adapted to DiT, MotionInversion suffers from severe overfitting, failing to edit the rider into an astronaut or modify the background. DreamBooth and DMT fail to capture accurate motion, often straightening originally curved trajectories.
In contrast, our method preserves fine‑grained motion patterns while supporting flexible, text‑driven edits.

\subsection{Quantitative Results}
Tab.~\ref{tab:quantitative_results} reports MTBench results. Our method, combined with HunyuanVideo, attains the highest motion fidelity and the most balanced trade‑off between edit and motion fidelity, with MOFT showing a marginal advantage in edit fidelity. Adaptations of MotionInversion, DreamBooth, and DMT to the DiT model consistently underperform across all metrics. As in prior studies, motion fidelity declines while edit fidelity improves with increasing difficulty, confirming the validity of our benchmark’s difficulty partitioning. Notably, our method achieves the top score at the medium level, driven by its substantial performance gains.

\subsection{Ablation Study}

\begin{table}[t]
\centering
\scalebox{0.75}{
\begin{tabular}{lccc}
\toprule
Method & Edit Fidelity & Temporal Consistency & Motion Fidelity \\
\midrule
LoRA             & 28.4 & 85.6 & \underline{80.4} \\
Conv3D           & 27.1 & 87.4 & 84.9 \\
Local Attn       & \underline{31.3} & \underline{88.2} & 73.1 \\
Temp Conv1D       & \textbf{31.6} & \textbf{90.4} & \textbf{85.6} \\
\bottomrule
\end{tabular}
}
\vspace{-5pt}
\caption{\textbf{Ablation on different motion learning methods.} The ablation experiments were run on a 30-video subset of MTBench, stratified to mirror the full benchmark’s distribution of difficulty levels and motion categories.}
\vspace{-10pt}
\label{tab:method_comparison}
\end{table}

\begin{table}[t]
\centering
\begin{tabular}{cc}
\begin{subtable}[t]{0.45\linewidth}
\centering
\scalebox{0.8}{%
\begin{tabular}{lccc}
\toprule
Percent & EF & TC & MF \\
\midrule
75\% & \underline{31.3} & 89.4 & 83.1 \\
65\% & \textbf{31.6} & \textbf{90.4} & \textbf{85.6} \\
55\% & 30.4 & \underline{90.3} & \underline{85.2} \\
\bottomrule
\end{tabular}%
}
\caption{Ablation study on the percentage of dropped layers.}
\label{tab:drop_percentage}
\end{subtable}
&
\begin{subtable}[t]{0.45\linewidth}
\centering
\scalebox{0.8}{%
\begin{tabular}{lccc}
\toprule
$\lambda_{TL}$ & EF & TC & MF \\
\midrule
1e-1 & \textbf{31.6} & \textbf{90.4} & \textbf{85.6} \\
1e-2 & \textbf{31.6} & \underline{90.2} & \underline{83.4} \\
w/o TL & \underline{31.2} & 89.6 & 82.3 \\
\bottomrule
\end{tabular}%
}
\caption{Ablation study on the weight of dense point tracking loss $\lambda_{TL}$.}
\label{tab:tracking_loss_weight_b}
\end{subtable}
\\[1ex] 
\begin{subtable}[t]{0.45\linewidth}
\centering
\scalebox{0.8}{%
\begin{tabular}{lccc}
\toprule
$k$ & EF & TC & MF \\
\midrule
3 & \textbf{31.6} & \textbf{90.4} & \textbf{85.6} \\
5 & \underline{31.1} & 89.3 & \underline{83.1} \\
7 & 30.7 & \underline{89.6} & 81.7 \\
\bottomrule
\end{tabular}%
}
\caption{Ablation study on the kernel size $k$ of shared temporal kernel.}
\label{tab:kernel_size}
\end{subtable}
&
\begin{subtable}[t]{0.45\linewidth}
\centering
\scalebox{0.8}{%
\begin{tabular}{lccc}
\toprule
$m$ & EF & TC & MF \\
\midrule
64  & \textbf{31.7} & \underline{90.1} & \underline{84.2} \\
128 & \underline{31.6} & \textbf{90.4} & \textbf{85.6} \\
256 & 30.9 & 89.9 & 82.6 \\
\bottomrule
\end{tabular}
}
\caption{Ablation study on the mid dim $m$ of shared temporal kernel.}
\label{tab:mid_dimension}
\end{subtable}
\end{tabular}
\vspace{-5pt}
\caption{\textbf{Ablation studies on the hyperparameters in DeT.}}
\vspace{-15pt}
\label{tab:combined_ablation}
\end{table}

\noindent
\textbf{Shared temporal kernel.}
We conduct the ablation study on the key design choices. Instead of using a shared temporal kernel, we consider alternative approaches through LoRA~\cite{lora}, Conv3D, and local attention~\cite{swin}. 
As shown in Tab.~\ref{tab:method_comparison}, our method achieves the best performance in edit and motion fidelity metrics.
Fig.~\ref{fig:temp_conv1d_ablation} shows that shared temporal kernel achieve decoupling while learning the motion. We attribute our success to effective temporal smoothness and an attention map-based motion learning method.
Additionally, we ablate the percentage of dropped layers, kernel size, and middle dimension in Tab.~\ref{tab:drop_percentage}, Tab.~\ref{tab:kernel_size}, and Tab.~\ref{tab:mid_dimension}, respectively. Quantitative results show that dropping 65\% of layers, using a kernel size of 3, and setting the middle dimension to 128 achieve the best performance.

\noindent
\textbf{Dense point tracking loss.}
In Tab.~\ref{tab:tracking_loss_weight_b}, we ablate the weight of the dense point tracking loss and find that $\lambda_{TL}$ = 1e-1 achieves the best results. 
Fig.~\ref{fig:tracking_loss_ablation_2} illustrates the cosine similarity between DiT features during training. The dense point tracking loss facilitates precise foreground tracking. 
Additionally, as shown in Fig.~\ref{fig:tracking_loss_ablation}, we conduct a qualitative comparison for removing the dense point tracking loss. The green and red boxes highlight the effect of dense point tracking loss, which enhances the motion consistency.

For additional ablation on the denoising loss weight $\lambda_{DL}$, please refer to the SM. Our qualitative and quantitative results show that reducing $\lambda_{DL}$ collapses background structures and increases edit fidelity, but simultaneously degrades motion fidelity.
\section{Conclusion}
\label{sec:conclusion}
We propose DeT, a tuning-based method that leverages DiT models for the motion transfer task. Our novel shared temporal kernel effectively decouples and learns the motion simultanously. We further introduce dense point tracking loss on latent features to further enhance the foreground motion's consistency. Additionally, we propose MTBench, a more challenging and general benchmark with a hybrid motion fidelity metric combining Fréchet distance with velocity similarity. Experimental results demonstrate DeT’s effectiveness in accurately transferring motion while maintaining strong text controllability.

\noindent
\textbf{Acknowledgement.} This work is supported by the National Key Research and Development Program of China (No. 2023YFC3807600).

{
    \small
    \bibliographystyle{ieeenat_fullname}
    \bibliography{main}
}
 
\clearpage
\setcounter{page}{1}
\maketitlesupplementary

\noindent
\textbf{Overview.}
\begin{itemize}
    \item \textbf{\cref{sec:supp-more-qualitative-results}.} More qualitative results.
    \item \textbf{\cref{sec:supp-compare}.} More qualitative comparison results.
    \item \textbf{\cref{sec:supp-more-quantitative-results}.} More quantitative results.
    \item \textbf{\cref{sec:supp-implementation-details}.} Implementation details of the experiments.
    \item \textbf{\cref{sec:supp-ablation-study}.} Additional analysis and ablation studies.
    \item \textbf{\cref{sec:supp-benchmark-details}.} Details of the MTBench.
    \item  \textbf{\cref{sec:additional_related_work}.} Additional related works.
    \item \textbf{\cref{sec:supp-limitations}.} Limitations and future work.
\end{itemize}

\section{More Qualitative Results}
\label{sec:supp-more-qualitative-results}

\begin{figure*}[t]
  \centering
  \includegraphics[width=1.0\textwidth]{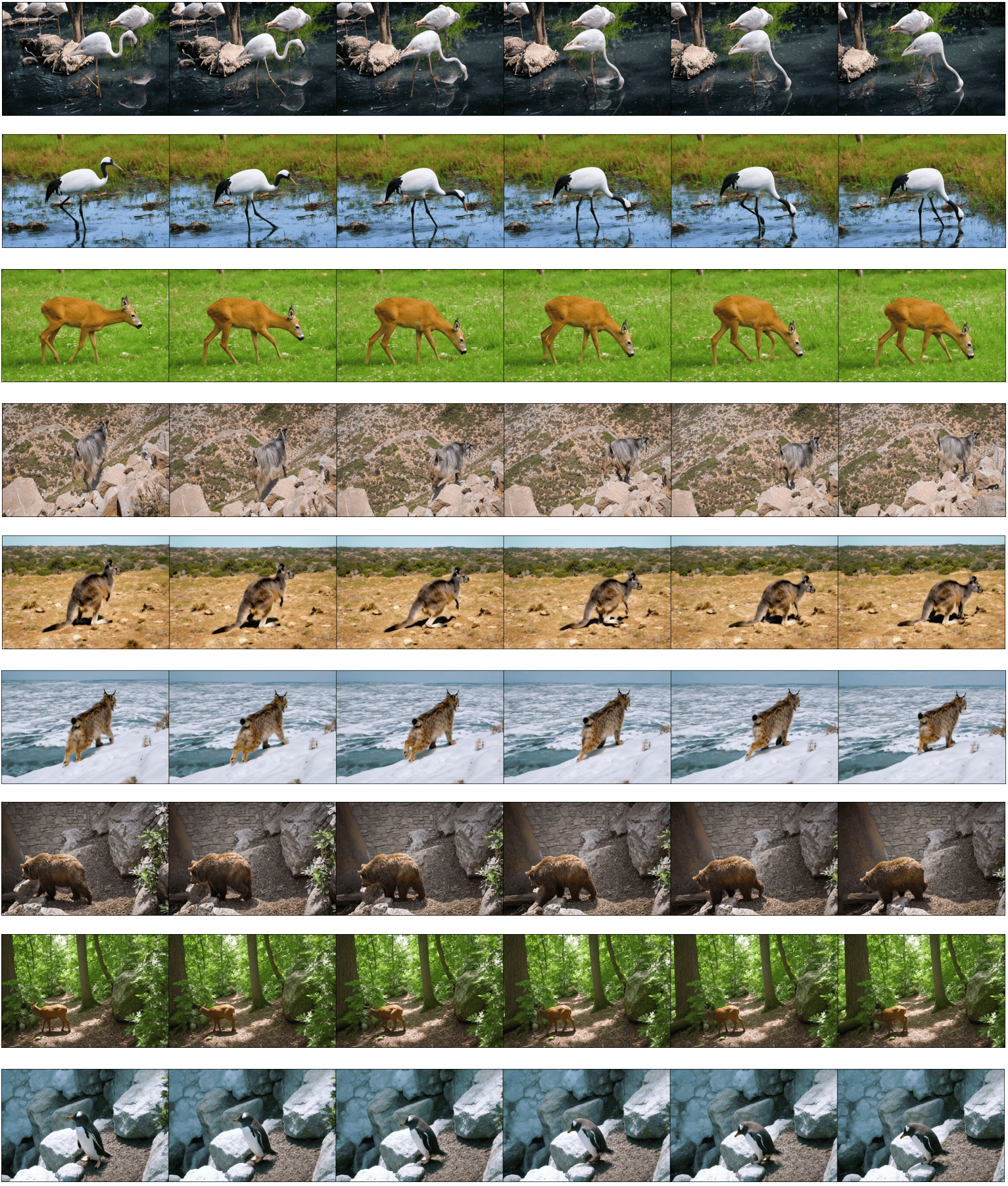}
  \caption{\textbf{More Qualitative Results.} There are three set of results. Each source video is combined with two newly generated videos on the bottom. Generated by DeT with Step-Video-T2V}
  \label{fig:supp_1}
\end{figure*}

\begin{figure*}[t]
  \centering
  \includegraphics[width=1.0\textwidth]{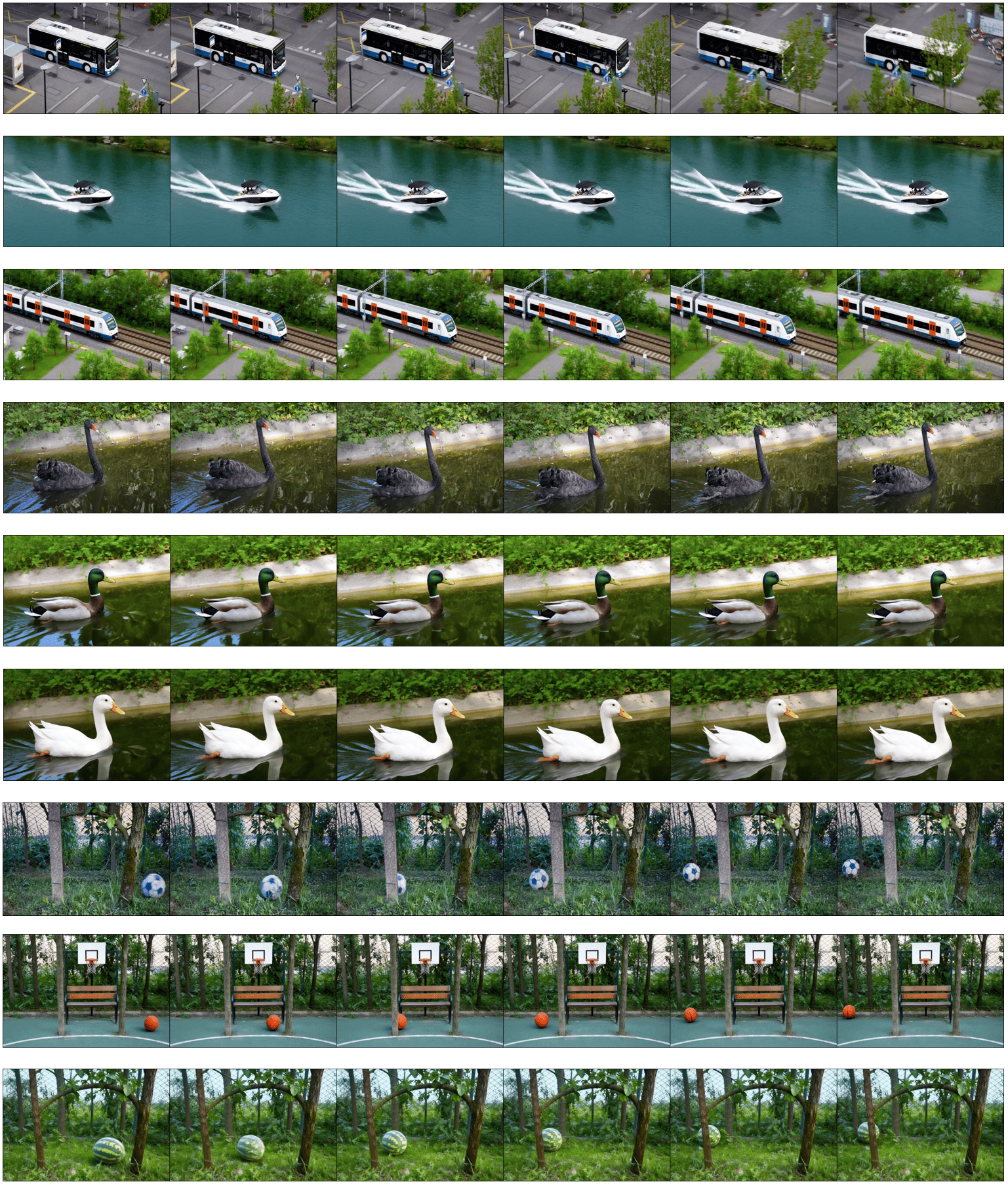}
  \caption{\textbf{More Qualitative Results.} There are three set of results. Each source video is combined with two newly generated videos on the bottom. Generated by DeT with HunyuanVideo}
  \label{fig:supp_2}
\end{figure*}

\begin{figure*}[t]
  \centering
  \includegraphics[width=1.0\textwidth]{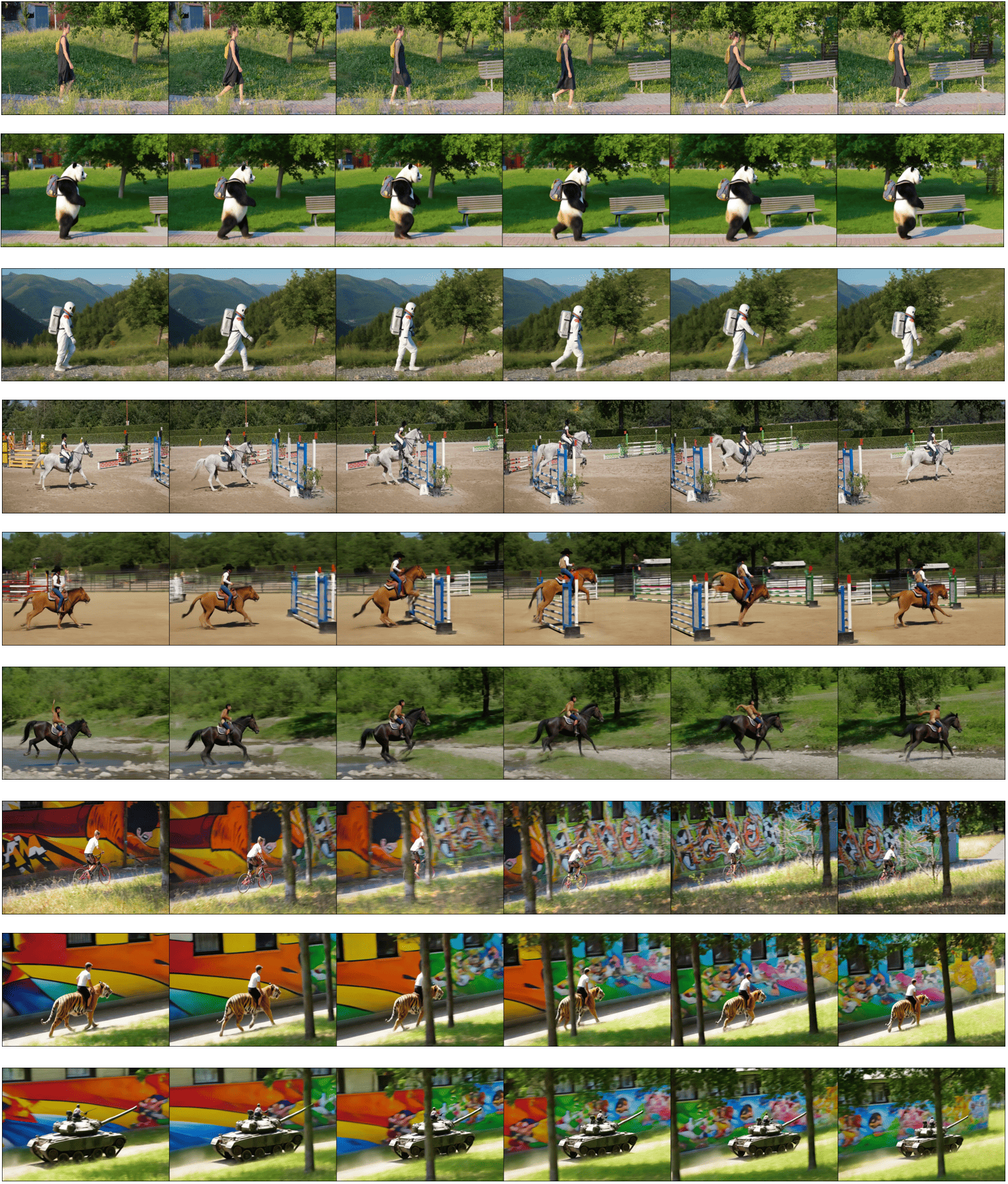}
  \caption{\textbf{More Qualitative Results.} There are three set of results. Each source video is combined with two newly generated videos on the bottom. Generated by DeT with Step-Video-T2V}
  \label{fig:supp_3}
\end{figure*}

We show more qualitative results in Fig.~\ref{fig:supp_1}, Fig.~\ref{fig:supp_2}, and Fig.~\ref{fig:supp_3}. Each source video is combined with two newly generated videos.

\section{More Qualitative Comparison Results}
\label{sec:supp-compare}

\begin{figure*}[t]
  \centering
  \includegraphics[width=1.0\textwidth]{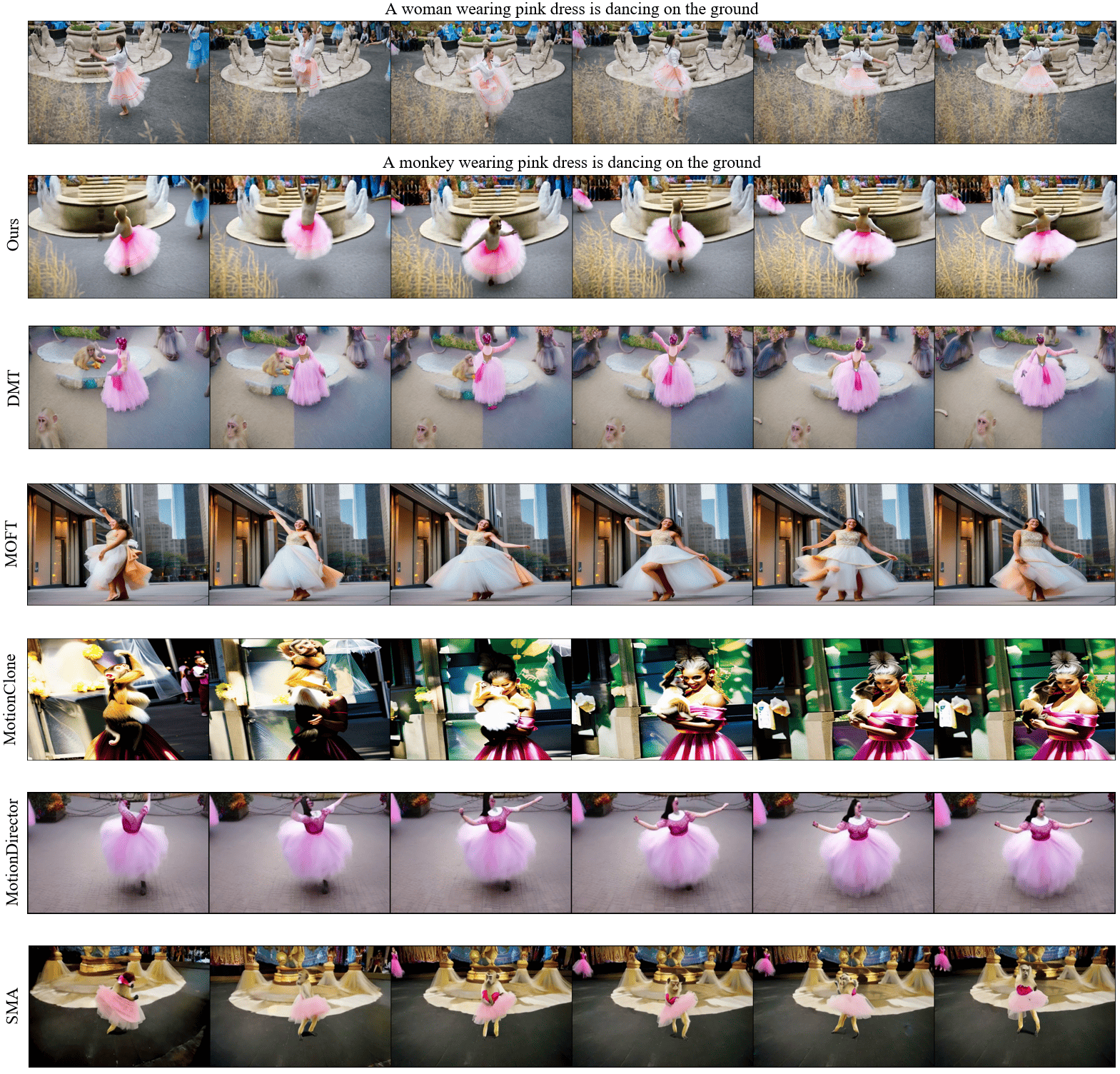}
  \caption{\textbf{More Qualitative Comparision Results.} Our result is generated by DeT with CogVideoX-5B}
  \label{fig:supp_4}
\end{figure*}

\begin{figure*}[t]
  \centering
  \includegraphics[width=1.0\textwidth]{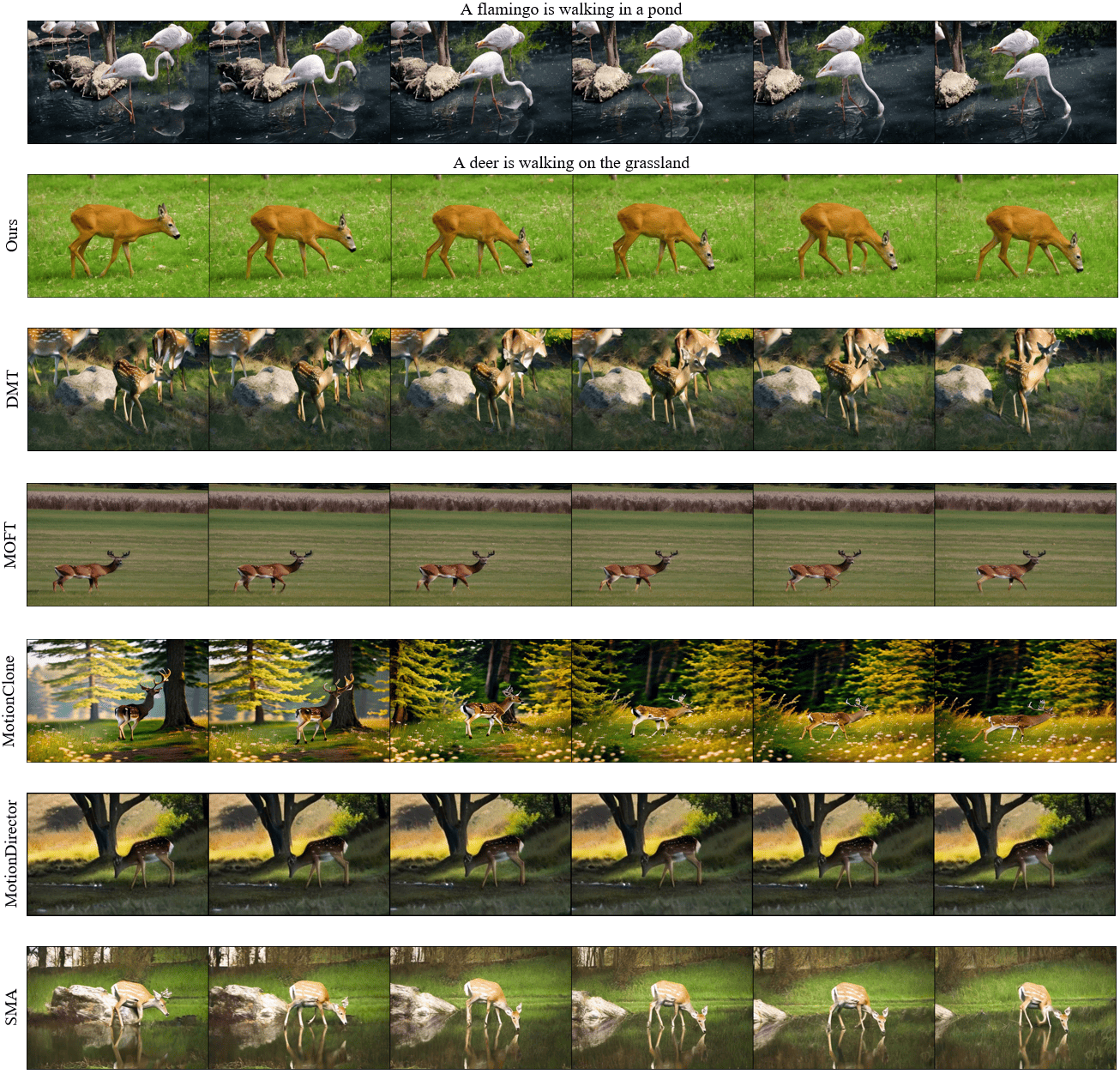}
  \caption{\textbf{More Qualitative Comparision Results.} Our result is generated by DeT with Step-Video-T2V}
  \label{fig:supp_5}
  \vspace{100pt}
\end{figure*}

\begin{figure*}[t]
  \centering
  \includegraphics[width=1.0\textwidth]{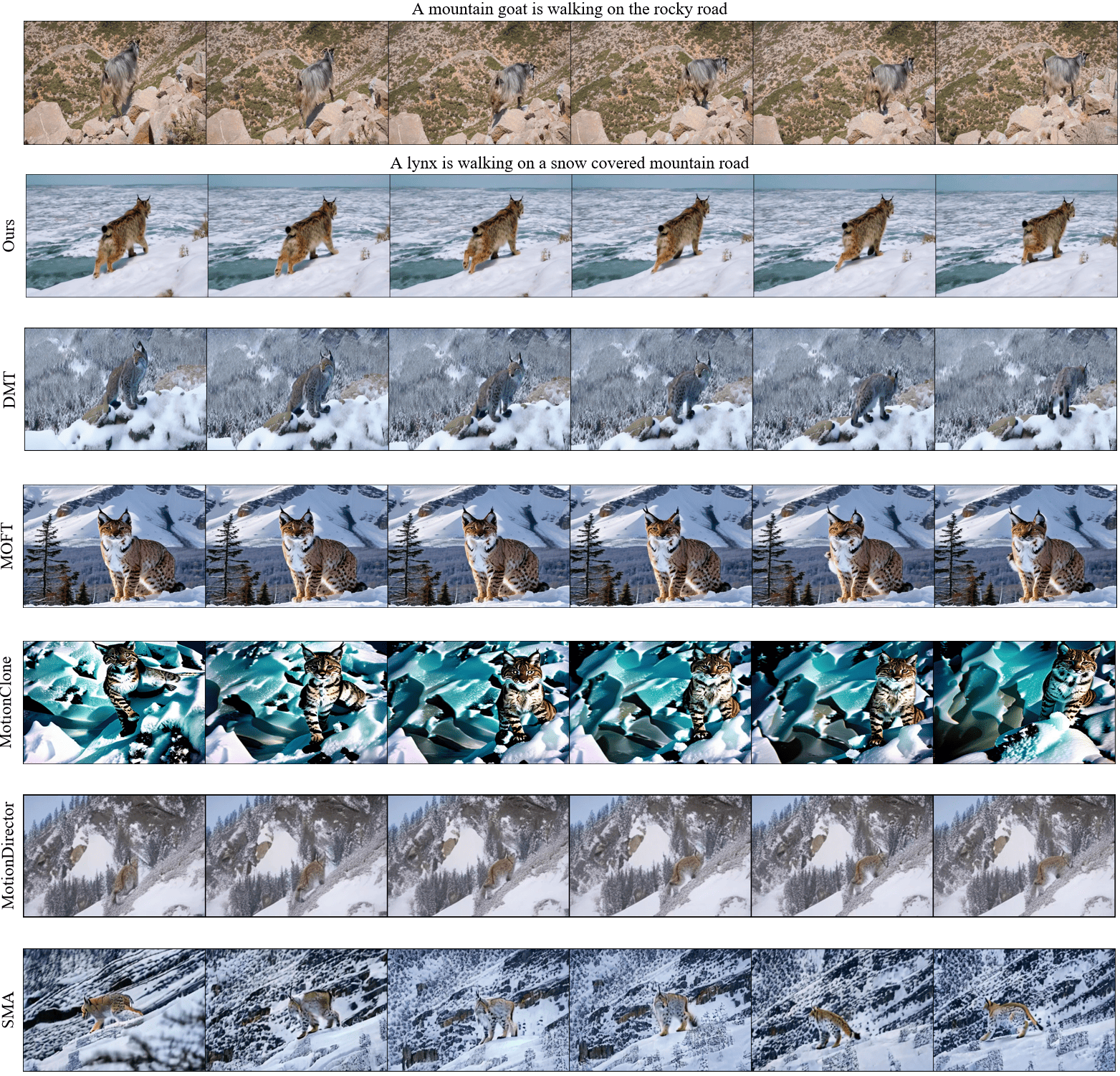}
  \caption{\textbf{More Qualitative Comparision Results.} Our result is generated by DeT with Step-Video-T2V}
  \label{fig:supp_6}
\end{figure*}

We present additional qualitative comparisons in Fig.~\ref{fig:supp_4}, Fig.~\ref{fig:supp_5}, and Fig.~\ref{fig:supp_6}. Our method accurately transfers complex motions while enabling flexible foreground and background control. These results demonstrate the generality of our motion transfer approach.

\section{More quantitative Results}
\label{sec:supp-more-quantitative-results}

\noindent
\textbf{V2VBench.}
We further evaluate our approach on V2VBench. As shown in Tab.~\ref{tab:V2VBench}, DeT demonstrates a clear advantage in both motion and video-text alignment—key aspects of the motion transfer task.

\noindent
\textbf{User Study.}
We conduct a user preference evaluation. As shown in Tab.~\ref{tab:user_study_metrics}, DeT demonstrates a clear advantage across all assessed aspects.

\section{Implementation Details}
\label{sec:supp-implementation-details}

\noindent
\textbf{HunyuanVideo implementation details.}
We train DeT on HunyuanVideo for 500 steps using the AdamW optimizer with a learning rate of 1e-5 and a weight decay of 1e-2. The learning rate is linearly warmed up over the first 100 steps. The loss weight parameters $\lambda_{DL}$ and $\lambda_{TL}$ are set to 1.0 and 0.1, respectively. For the motion module, the mid-dimension is set to 128, and the kernel size is configured to 5. 
During training, we integrate our shared temporal kernel into all DiT blocks. During inference, we remove it from the last 40 blocks (66\%) -  all of the single-stream DiT blocks. We perform 30 steps denoising using the Flow Matching scheduler. The generated videos have a resolution of 49 × 512 × 768. The training process takes approximately 1.5 hours on a single NVIDIA A100 80GB GPU.

\noindent
\textbf{Step-Video-T2V implementation details.}
We train DeT on Step-Video-T2V for 500 steps using the AdamW optimizer with a learning rate of 2e-5 and a weight decay of 1e-2. The learning rate is linearly warmed up over the first 100 steps. The loss weight parameters \(\lambda_1\) and \(\lambda_2\) are set to 1.0 and 0.1, respectively.  
For the shared temporal kernel, the mid-dimension is set to 256, and the kernel size is configured to 5. During training, we integrate the shared temporal kernel into all DiT blocks. During inference, we remove it from the last 32 blocks (66\%). We perform 50-step denoising using the Flow Matching (FM) scheduler with a classifier-free guidance scale of 9.0. The generated videos have a resolution of \(49 \times 512 \times 768\). The entire training process takes approximately two hours on a single NVIDIA A100 80GB GPU.

\begin{table}[t]
  \centering
  \caption{V2VBench results (higher is better).}
  \vspace{-10pt}
  \resizebox{1.0\linewidth}{!}{%
  \begin{tabular}{lrrrrrrrr}
    \toprule
    \textbf{Method} &
    \textbf{Motion} &
    \textbf{Video} &
    \textbf{Frames} &
    \textbf{Object} &
    \textbf{Semantic} &
    \textbf{Video} &
    \textbf{Frames} &
    \textbf{Frames} \\[-0.25em]
    &
    \textbf{Align $\uparrow$} &
    \textbf{Txt $\uparrow$} &
    \textbf{Txt $\uparrow$} &
    \textbf{Cons. $\uparrow$} &
    \textbf{Cons. $\uparrow$} &
    \textbf{Qual. $\uparrow$} &
    \textbf{Qual. $\uparrow$} &
    \textbf{Pick $\uparrow$} \\
    \midrule
    MotionDirector & $-3.09$ & $20.92$ & \textbf{27.85} & $0.94$ & $0.95$ & $0.62$ & $4.98$ & $0.26$ \\
    TokenFlow      & $-1.57$ & $20.76$ & $27.52$ & $0.95$ & $0.94$ & \textbf{0.72} & \textbf{5.07} & $0.25$ \\
    DeT (Ours)     & \textbf{-0.83} & \textbf{31.50} & $27.84$ & \textbf{0.96} & \textbf{0.97} & $0.63$ & $4.99$ & \textbf{0.29} \\
    \bottomrule
  \end{tabular}}
  \vspace{-20pt}
  \label{tab:V2VBench}
\end{table}

\noindent
\textbf{Baselines implementation details.}
We implement MotionInversion~\cite{motioninversion} on CogVideoX-5B~\cite{cogvideox} by injecting temporal embeddings \(E_t \in \mathbb{R}^{t \times c}\) and spatial embeddings \(E_s \in \mathbb{R}^{h \times w \times c}\) into the 3D full attention mechanism. Specifically, we inject \(E_t\) into the query and key, while \(E_s\) is applied to the value.  
We train MotionInversion for 500 steps using the AdamW optimizer with a learning rate of 1e-3 and a weight decay of 1e-2, injecting embeddings into all DiT blocks. For DreamBooth, we use the LoRA variant, setting the LoRA rank to 128 and injecting LoRA weights into the query, key, value, and output linear layers. We train with a learning rate of 1e-4 and a weight decay of 1e-2.  
For DMT, we apply DDIM inversion to each video, which takes approximately one hour. During inference, we extract features from the 28th DiT block and compute SMM~\cite{DMT} via spatial pooling and first-order differencing. The optimization learning rate is set to 1e-2, and we optimize the latents for 20 steps. 
For other U-Net-based methods, we use their official implementations and settings.

\section{Additional Analysis and Ablation Study}
\label{sec:supp-ablation-study}

\begin{figure}[t]
  \centering
  \includegraphics[width=0.45\textwidth]{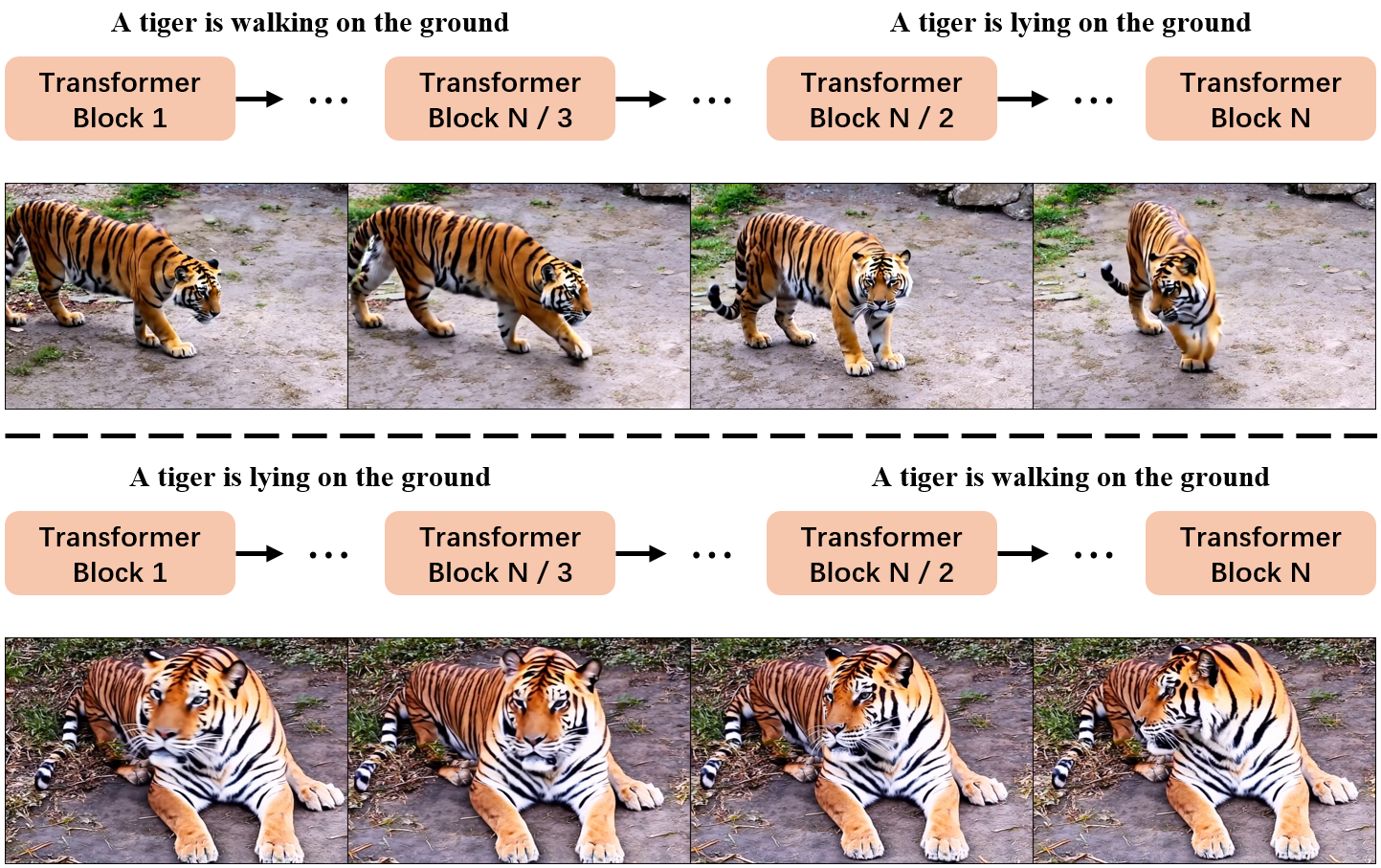}
  \caption{\textbf{Analysis of dropping layer strategy.} The generation results of DiT models rely more on features from earlier layers. The tiger's motion is determined by the text prompt in these early layers.}
  \label{fig:drop_analyze_1}
\end{figure}

\begin{figure}[t]
  \centering
  \includegraphics[width=0.45\textwidth]{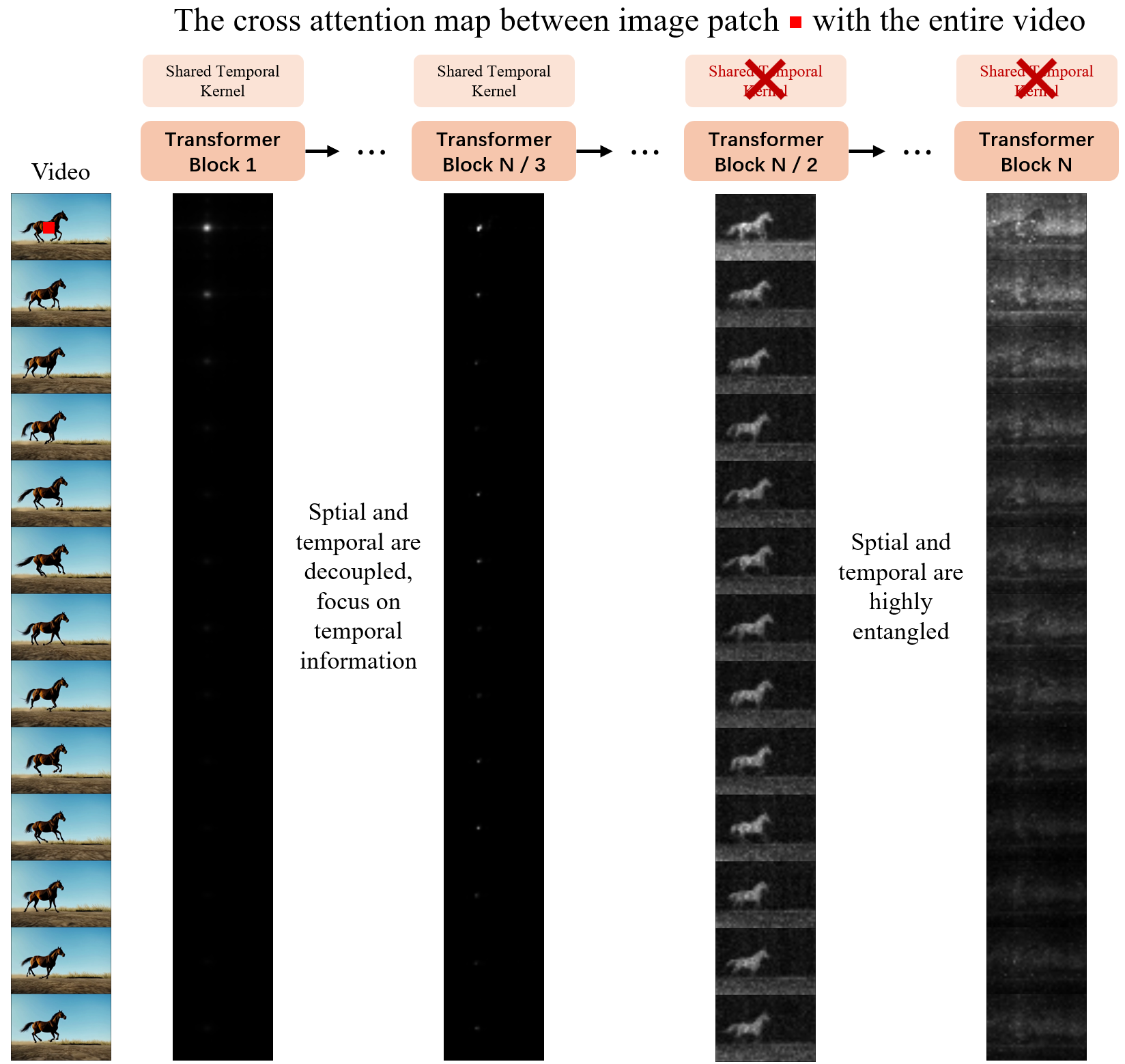}
  \caption{\textbf{Anaylsis of dropping layer strategy.} The attention map also supports our dropping layer strategy. As the later layers' attention maps show entangled spatial and temporal information, decoupling becomes ineffective. To improve decoupling, we remove the shared temporal kernel in these layers.}
  \label{fig:drop_analyze_2}
\end{figure}

\subsection{Analysis for drop layers}
During inference, we drop the shared temporal kernel in the last 65\% of DiT blocks. Experimental results show that this dropping strategy enhances text controllability. We attribute this to the fact that the DiT model relies more on the earlier layers when generating videos, as shown in Fig.~\ref{fig:drop_analyze_1}. This suggests that the shared temporal kernel inserted in the later layers may introduce redundant learnable parameters that capture information unrelated to motion.

Furthermore, the visualization of the attention map in Fig.~\ref{fig:drop_analyze_2} reveals that in the earlier layers, the DiT model exhibits a mechanism similar to temporal self-attention, whereas the later layers resemble spatiotemporal attention. Spatiotemporal attention is less effective at decoupling motion from appearance, which further supports the decision to drop the temporal kernel in the later layers. Additionally, we observe that applying this layer-dropping strategy consistently improves performance across all three DiT models.

\begin{figure}[t]
  \centering
  \includegraphics[width=0.47\textwidth]{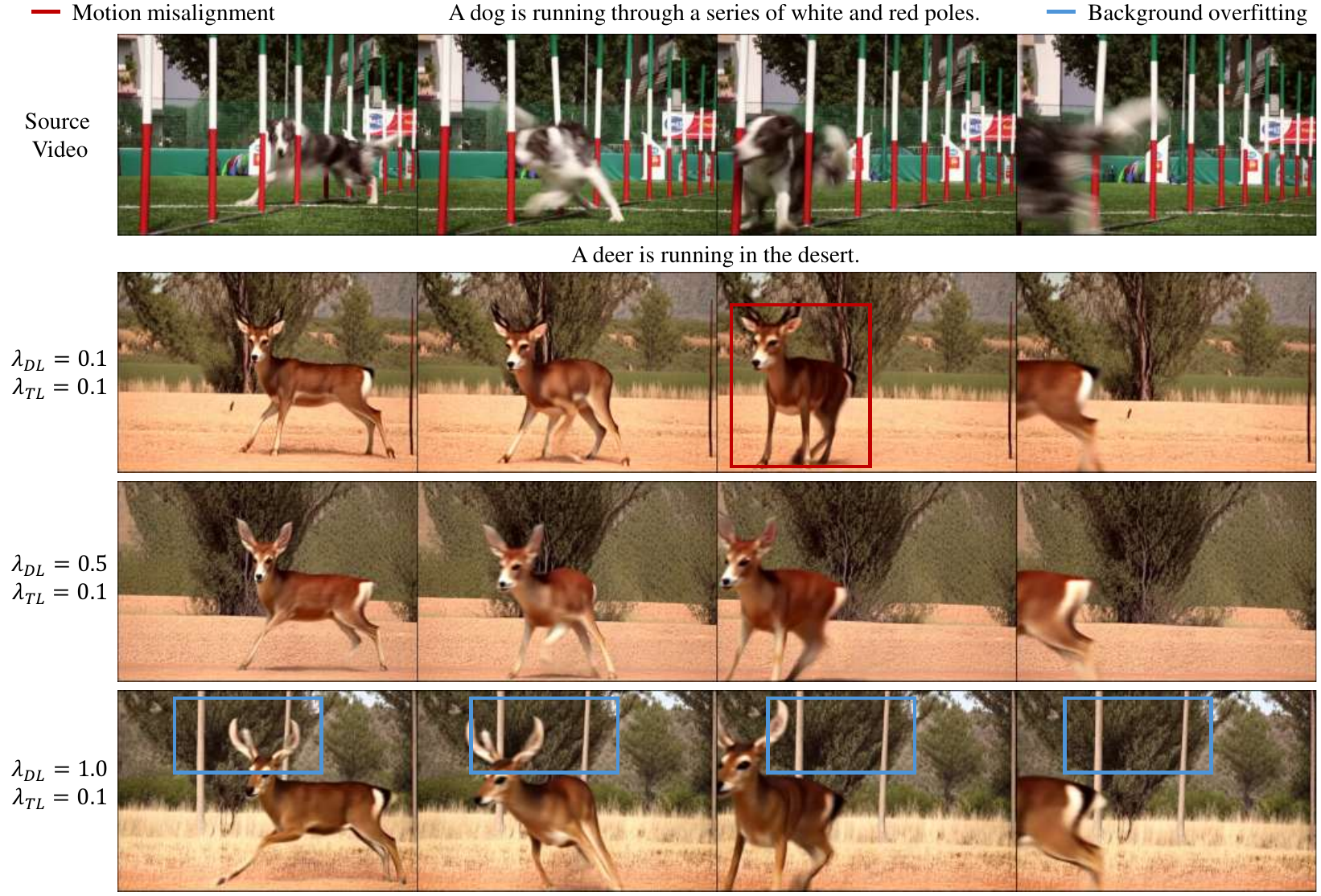}
  \vspace{-10pt}
  \caption{Ablation of the denoising loss weight $\lambda_{DL}$.}
  \vspace{-10pt}
  \label{fig:ablation_dl}
\end{figure}

\subsection{Analysis for denoising loss weight}
Our ablation on the denoising loss weight $\lambda_{DL}$ (Fig.~\ref{fig:ablation_dl}, Tab.~\ref{tab:ablation_dl}), shows that reducing $\lambda_{DL}$ collapses background structure and boosts edit fidelity, but simultaneously degrades motion fidelity.

\subsection{DeT with HunyuanVideo}

Table~\ref{tab:hunyuan_combined_ablation} presents comprehensive ablation studies conducted on various hyperparameters within the DeT framework with HunyuanVideo. 

\begin{table}[t]
\centering
\begin{tabular}{cc}
\begin{subtable}[t]{0.45\linewidth}
\centering
\scalebox{0.8}{%
\begin{tabular}{lccc}
\toprule
Percent & EF & TC & MF \\
\midrule
75\% & \underline{31.2} & \underline{89.1} & 83.4 \\
65\% & \textbf{31.9} & \textbf{91.9} & \textbf{85.9} \\
55\% & 30.7 & 89.0 & \underline{85.5} \\
\bottomrule
\end{tabular}%
}
\caption{Ablation study on the percentage of dropped layers.}
\label{tab:hunyuan_drop_percentage}
\end{subtable}
&
\begin{subtable}[t]{0.45\linewidth}
\centering
\scalebox{0.8}{%
\begin{tabular}{lccc}
\toprule
$\lambda_{TL}$ & EF & TC & MF \\
\midrule
1e-1 & \textbf{31.9} & \textbf{91.9} & \textbf{85.9} \\
1e-2 & \underline{31.7} & \underline{90.1} & \underline{84.4} \\
w/o TL & 31.6 & 89.4 & 83.1 \\
\bottomrule
\end{tabular}%
}
\caption{Ablation study on the weight of dense point tracking loss $\lambda_{TL}$.}
\label{tab:hunyuan_tracking_loss_weight}
\end{subtable}
\\[1ex] 
\begin{subtable}[t]{0.45\linewidth}
\centering
\scalebox{0.8}{%
\begin{tabular}{lccc}
\toprule
$k$ & EF & TC & MF \\
\midrule
3 & \underline{31.4} & \underline{90.6} & 85.1 \\
5 & \textbf{31.9} & \textbf{91.9} & \textbf{85.9} \\
7 & 30.7 & 89.6 & \underline{85.7} \\
\bottomrule
\end{tabular}%
}
\caption{Ablation study on the kernel size $k$ of shared temporal kernel.}
\label{tab:hunyuan_kernel_size}
\end{subtable}
&
\begin{subtable}[t]{0.45\linewidth}
\centering
\scalebox{0.8}{%
\begin{tabular}{lccc}
\toprule
$m$ & EF & TC & MF \\
\midrule
64  & 30.2 & \underline{90.1} & 83.2 \\
128 & \textbf{31.9} & \textbf{91.9} & \textbf{85.9} \\
256 & \underline{30.9} & 89.9 & \underline{84.6} \\
\bottomrule
\end{tabular}
}
\caption{Ablation study on the mid dim $m$ of shared temporal kernel.}
\label{tab:hunyuan_mid_dimension}
\end{subtable}
\end{tabular}
\vspace{-5pt}
\caption{Ablation studies on the hyperparameters in DeT with HunyuanVideo.}
\vspace{-10pt}
\label{tab:hunyuan_combined_ablation}
\end{table}

\noindent
\textbf{Drop layers.}
Fig.~\ref{tab:hunyuan_combined_ablation} (a) investigates the impact of varying the percentage of dropped layers. The best performance is achieved at 65\%, as indicated by the highest scores. This result highlights the importance of a balanced dropout rate in optimizing the model's motion transfer capabilities.

\noindent
\textbf{Dense point tracking loss.}
Fig.~\ref{tab:hunyuan_tracking_loss_weight}, we ablate the dense point tracking loss weight for DeT with HunyuanVideo. We find that $\lambda_{TL}=\text{1e-1}$ achieves the best results. 

\noindent
\textbf{Temporal kernel size.}
Fig.~\ref{tab:hunyuan_combined_ablation} (c) ablates the kernel size, with results demonstrating that a kernel size of 5 offers optimal performance. This suggests that the receptive field provided by $k=5$ strikes the best balance between capturing local temporal dependencies and maintaining computational efficiency.

\noindent
\textbf{Mid dimension}
Fig.~\ref{tab:hunyuan_combined_ablation} (d) explores the mid dimension parameter, showing that a dimension of 128 is most effective.

\subsection{DeT with Step-Video-T2V}

Table~\ref{tab:hunyuan_combined_ablation} presents comprehensive ablation studies conducted on various hyperparameters within the DeT with Step-Video-T2V. 

\begin{table}[t]
\centering
\begin{tabular}{cc}
\begin{subtable}[t]{0.45\linewidth}
\centering
\scalebox{0.8}{%
\begin{tabular}{lccc}
\toprule
Percent & EF & TC & MF \\
\midrule
75\% & \underline{31.1} & 88.3 & 82.3 \\
65\% & \textbf{31.4} & \textbf{91.6} & \textbf{85.8} \\
55\% & 30.2 & \underline{89.0} & \underline{84.2} \\
\bottomrule
\end{tabular}%
}
\caption{Ablation study on the percentage of dropped layers.}
\label{tab:step_drop_percentage}
\end{subtable}
&
\begin{subtable}[t]{0.45\linewidth}
\centering
\scalebox{0.8}{%
\begin{tabular}{lccc}
\toprule
$\lambda_{TL}$ & EF & TC & MF \\
\midrule
1e-1 & \textbf{31.4} & \textbf{91.6} & \textbf{85.8} \\
1e-2 & \underline{30.5} & \underline{90.5} & \underline{84.9} \\
w/o TL & 31.4 & 88.5 & 82.9 \\
\bottomrule
\end{tabular}%
}
\caption{Ablation study on the weight of dense point tracking loss $\lambda_{TL}$.}
\label{tab:step_tracking_loss_weight}
\end{subtable}
\\[1ex] 
\begin{subtable}[t]{0.45\linewidth}
\centering
\scalebox{0.8}{%
\begin{tabular}{lccc}
\toprule
$k$ & EF & TC & MF \\
\midrule
3 & \underline{31.1} & \underline{90.1} & 85.3 \\
5 & \textbf{31.4} & \textbf{91.6} & \textbf{85.8} \\
7 & 30.4 & 88.6 & \underline{85.4} \\
\bottomrule
\end{tabular}%
}
\caption{Ablation study on the kernel size $k$ of shared temporal kernel.}
\label{tab:step_kernel_size}
\end{subtable}
&
\begin{subtable}[t]{0.45\linewidth}
\centering
\scalebox{0.8}{%
\begin{tabular}{lccc}
\toprule
$m$ & EF & TC & MF \\
\midrule
64  & 31.2 & 90.5 & 83.7 \\
128 & \textbf{31.4} & \underline{91.2} & \underline{84.5} \\
256 & \textbf{31.4} & \textbf{91.6} & \textbf{85.8} \\
\bottomrule
\end{tabular}
}
\caption{Ablation study on the mid dim $m$ of shared temporal kernel.}
\label{tab:step_mid_dimension}
\end{subtable}
\end{tabular}
\vspace{-5pt}
\caption{Ablation studies on the hyperparameters in DeT with Step-Video-T2V.}
\vspace{-10pt}
\label{tab:step_combined_ablation}
\end{table}

\noindent
\textbf{Drop layers.}
Table~\ref{tab:step_drop_percentage} presents the ablation study on the percentage of dropped layers. The best performance is achieved at a 65\% drop rate.

\noindent
\textbf{Dense point tracking loss.}
In Table~\ref{tab:step_tracking_loss_weight}, the impact of the weight $\lambda_{TL}$ for the dense point tracking loss is evaluated. The results show that setting $\lambda_{TL}$ to 1e-1 yields the highest performance across all metrics.

\noindent
\textbf{Temporal kernel size.}
Table~\ref{tab:step_kernel_size} investigates the effect of different kernel sizes for the shared temporal convolution. The experiment reveals that a kernel size of 5 produces the best results, indicating an optimal balance.

\noindent
\textbf{Mid dimension}
Table~\ref{tab:step_mid_dimension} explores the influence of the mid dimension $m$ of the shared temporal kernel. The performance consistently improves with an increase in $m$, reaching optimal values when $m$ is set to 256.

\begin{table}[t]
  \centering
  \renewcommand{\arraystretch}{0.95}
  \setlength{\tabcolsep}{3pt}
  \scriptsize
  \begin{minipage}[t]{0.22\textwidth}
    \centering
    \captionof{table}{Ablation on $\lambda_{DL}$.}
    \vspace{-10pt}
    \begin{tabular}{lccc}
      \toprule
      $\lambda_{\text{DL}}$ & EF & TC & MF \\
      \midrule
      0.1 & \textbf{32.1} & \underline{90.3} & 85.2 \\
      0.5 & \underline{31.7} & \textbf{90.4} & \underline{85.4} \\
      1.0 & 31.6 & \textbf{90.4} & \textbf{85.6} \\
      \bottomrule
    \end{tabular}
    \label{tab:ablation_dl}
  \end{minipage}
  \hspace{1mm}
  \begin{minipage}[t]{0.22\textwidth}
    \centering
    \captionof{table}{User study results.}
    \vspace{-10pt}
    \begin{tabular}{lccc}
      \toprule
      Method & EF & TC & MF \\
      \midrule
      MD  & \underline{16.7}\% & 20\% & 5\% \\
      MI  & 10\% & \underline{23}\% & \underline{40}\% \\
      DeT & \textbf{73.3}\% & \textbf{56}\% & \textbf{55}\% \\
      \bottomrule
    \end{tabular}
    \label{tab:user_study_metrics}
  \end{minipage}
  \vspace{-10pt}
\end{table}

\section{MTBench Details}
\label{sec:supp-benchmark-details}

\begin{figure}[t]
  \centering
  \includegraphics[width=0.45\textwidth]{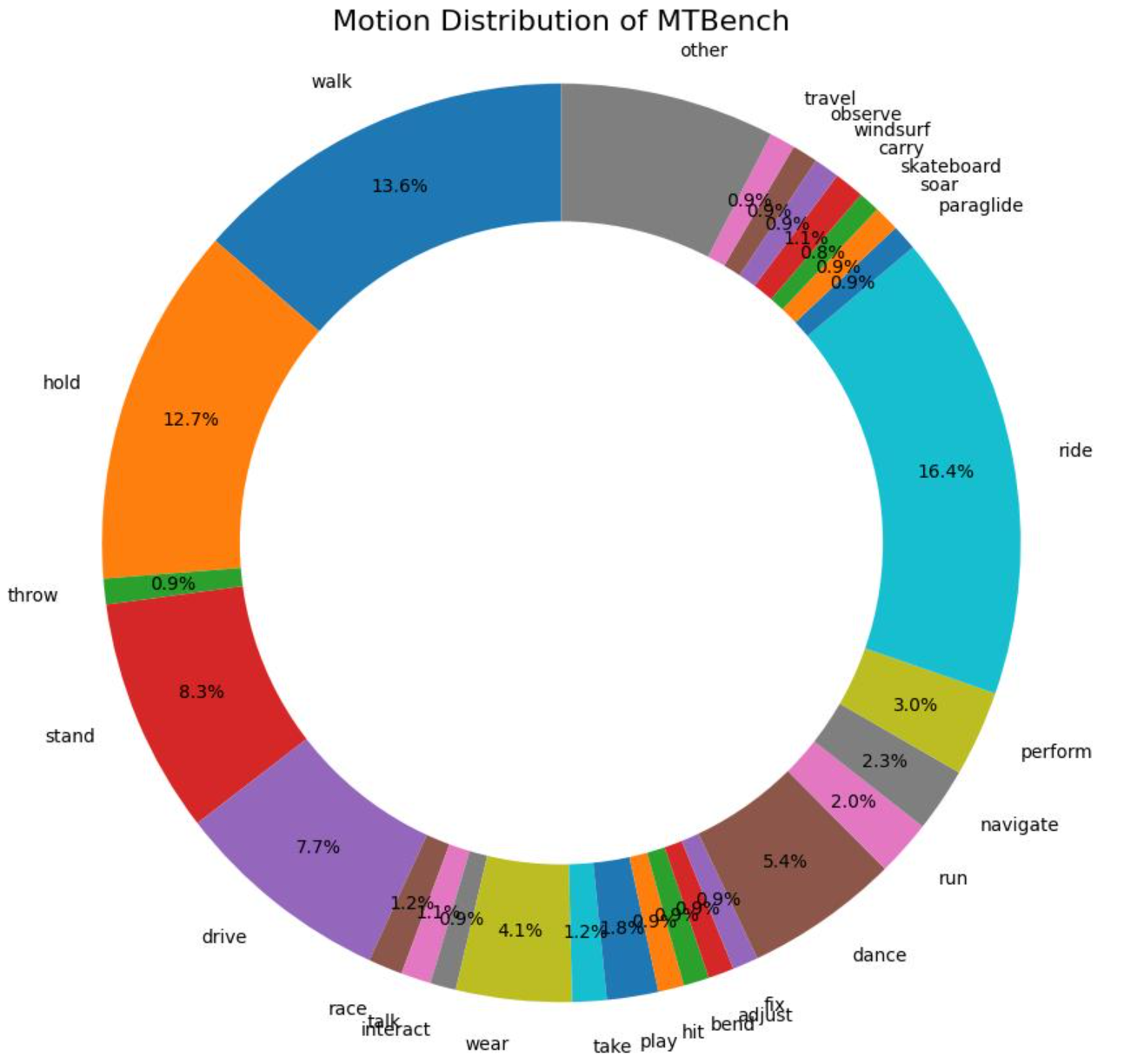}
  \caption{\textbf{Motion Distribution of MTBench.}}
  \label{fig:mtbench_1}
\end{figure}

\begin{figure}[t]
  \centering
  \includegraphics[width=0.45\textwidth]{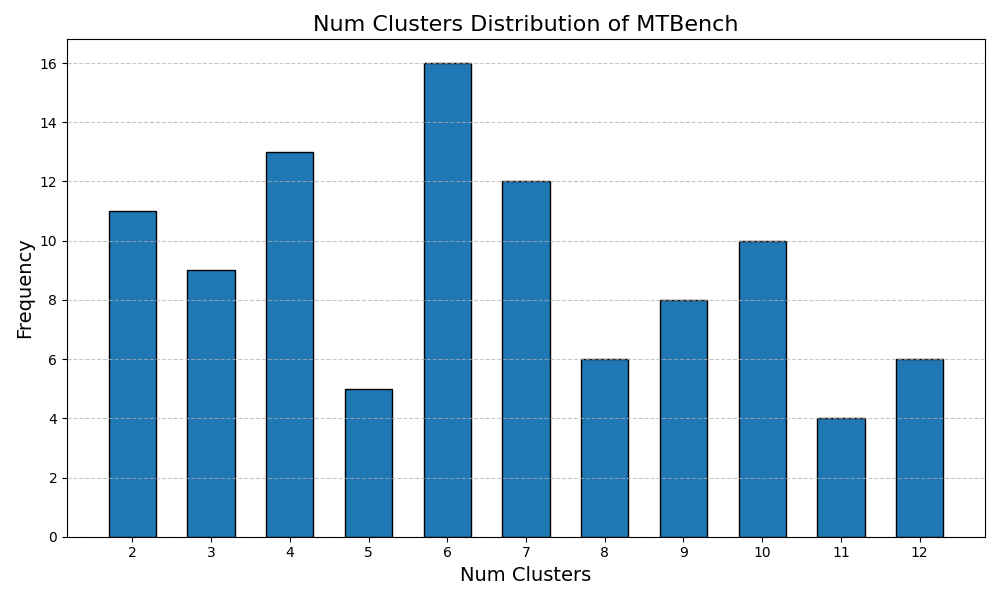}
  \caption{\textbf{Num Clusters (Difficulty) Distribution of MTBench.}}
  \label{fig:mtbench_2}
\end{figure}

\begin{figure}[t]
  \centering
  \includegraphics[width=0.45\textwidth]{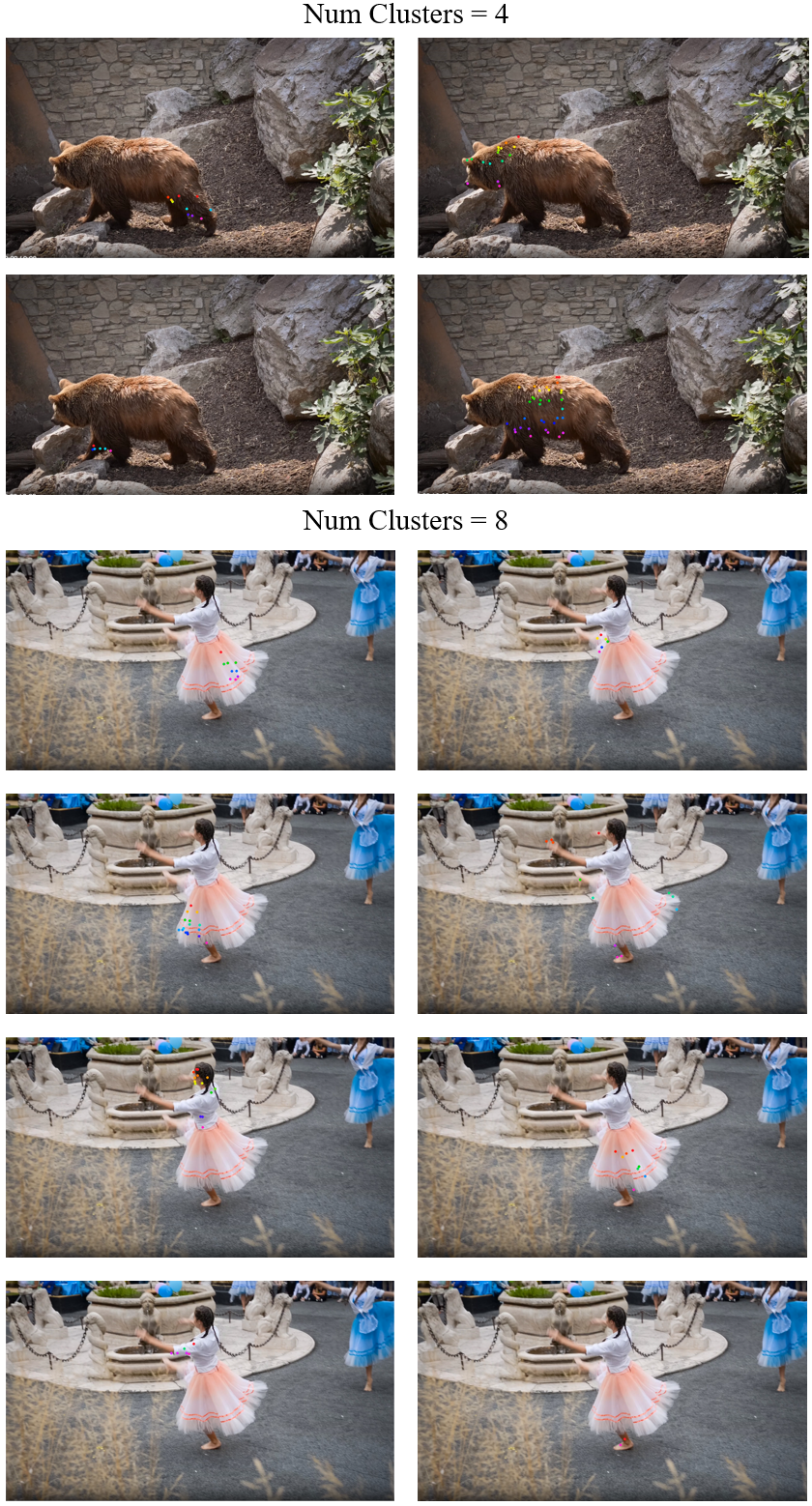}
  \caption{\textbf{Visualization of Clusters.} }
  \label{fig:mtbench_3}
\end{figure}

MTBench consists of 48 diverse motion categories specifically curated for evaluating motion transfer tasks. We visualize the motion distribution and the number of clusters in MTBench in Fig.~\ref{fig:mtbench_1} and Fig.~\ref{fig:mtbench_2}, respectively. Each category is annotated with its occurrence frequency. By incorporating both high-frequency and rare motions, MTBench provides a rigorous assessment of a motion transfer method’s generalization and robustness across a wide range of motion complexities.

Additionally, we visualize the trajectory clusters in Fig.~\ref{fig:mtbench_3}. The center of each cluster corresponds to a fundamental movement element of the foreground, such as a bear’s limb. This suggests that aligning the cluster centers of trajectories between the generated and source videos could serve as a potential decoupling method.

\section{Additional Related Works}
\label{sec:additional_related_work}

\noindent
\textbf{Controllable video generaion.} 
To better meet user demands, previous works~\cite{motionctrl, videocomposer, moviegen, alignyourlatents} have incorporated additional control signals into the video generation process, including control over the first frame~\cite{svd}, motion trajectories~\cite{dragnuwa}, object regions~\cite{motionbooth}, and object identity~\cite{dreamvideo, relationbooth}.

Specifically, trajectory-controlled methods guide subject movement through per-frame coordinates but lack fine-grained motion control. Region-controlled methods use bounding boxes to constrain subject positions in each frame, yet they still struggle to generate complex motions. Additionally, recent works~\cite{mvideo} employ video masks to regulate motion. While masks effectively guide motion generation, they also restrict subject appearance, reducing text controllability.

In this work, we propose a tuning-based motion transfer method for DiT models, where the generated videos follow the motion of the source video while maintaining strong text controllability.

\section{Limitations and Future Work}
\label{sec:supp-limitations}

The success of DeT relies on two key assumptions.

First, the foreground and background DiT features must be separable in high-dimensional space. If this condition holds, smoothing along the temporal dimension can help the model better distinguish between foreground and background. However, if these features are not separable in high-dimensional space or remain indistinguishable within a given temporal window defined by the kernel size, then temporal smoothing alone will be ineffective. In such cases, the model may still memorize background appearance, leading to overfitting.

Second, our assumption regarding the dense point tracking loss is that the key regions of foreground motion are already present in the first frame. This enables CoTracker to track foreground motion throughout the sequence. However, if critical motion-related parts are absent in the first frame—for example, a hidden hand—then the effectiveness of the dense point tracking loss may be reduced.

In the future, we will continue to enhance the model's ability to decouple and learn motion. While the shared temporal kernel handles both tasks simultaneously, designing dedicated modules for decoupling and motion learning separately may further improve overall performance.

\end{document}